%% file: main.tex
\documentclass{article} 
\ifdefined\pdfsuppressptexinfo \pdfsuppressptexinfo=-1 \fi  
\usepackage{arxiv_preprint,times}

\input{math_commands.tex}

\usepackage{hyperref}
\usepackage{url}

\usepackage{amsmath,amsfonts,bm}

\usepackage{array}
\usepackage{url}
\usepackage{booktabs}       
\usepackage{amsfonts}       
\usepackage{nicefrac}       
\usepackage{microtype}      
\usepackage{xcolor}         
\usepackage{natbib}         
\usepackage{makecell}       
\usepackage{multirow}       
\usepackage{multicol}
\usepackage{amsmath}        
\usepackage{graphicx}       
\usepackage{subcaption}  
\usepackage{color}
\usepackage[normalem]{ulem}
\usepackage{tikz}
\usepackage{colortbl}
\usepackage{pifont}
\usepackage{xspace}
\usepackage{enumitem}
\usepackage{appendix}
\usepackage{mathtools}
\usepackage{wrapfig}
\usepackage{hyperref}
\usepackage{algorithm}
\usepackage{algpseudocode}
\usepackage{setspace}
\usepackage{varwidth}
\usepackage{amssymb}
\usepackage{wrapfig}
\usepackage{booktabs}
\usepackage{multirow}
\usepackage{makecell}
\usepackage{colortbl}

\makeatletter
\@ifundefined{insert@pcolumn}{%
	\let\insert@pcolumn\insert@column
}{}
\makeatother

\newtheorem{Theorem}{Theorem}

\def\l{\left}
\def\r{\right}
\def\({\l(}
\def\){\r)}
\def\[{\l[}
\def\]{\r]}

\title{Flash EQ-Linear: Accelerating Equivariant Linear Layers via Group-wise Discrete Fourier Transform}

\author{
    \textbf{Zhongchen Zhao$^{1}$,
    Jixin Wang$^{1}$,
    Qi Xie$^{1,}$\thanks{Corresponding author.},
    Hui Lin$^{1}$,}\\[3pt]
    \textbf{Lei Zhang$^{2,3}$,
    Deyu Meng$^{1}$,
    Zongben Xu$^{1}$}\\[6pt]
    $^{1}$Xi'an Jiaotong University\\
    $^{2}$The Hong Kong Polytechnic University \qquad
    $^{3}$OPPO Research\\[3pt]
    \texttt{zhongchenzhao@stu.xjtu.edu.cn, xie.qi@mail.xjtu.edu.cn}
}
\date{}

\hypersetup{
    hidelinks,
    pdftitle={Flash EQ-Linear: Accelerating Equivariant Linear Layers via Group-wise Discrete Fourier Transform},
    pdfauthor={Zhongchen Zhao, Jixin Wang, Qi Xie, Hui Lin, Lei Zhang, Deyu Meng, Zongben Xu},
    pdfsubject={arXiv preprint}
}

\begin{document}

\maketitle

\begin{abstract}
	Equivariant networks embed geometric symmetries as structural priors through weight sharing, achieving remarkable parameter efficiency across vision tasks. However, this parameter efficiency does not translate into compute efficiency: most existing implementations unroll the structured weights into dense matrices and dispatch them to generic dense kernels, so an equivariant layer costs no fewer MACs than its non-equivariant counterpart. In this paper, we observe that the equivariant linear (EQ-Linear) layer---the most fundamental and frequently used module in modern equivariant architectures---is essentially a circular convolution along the group dimension composed with a linear transform along the channel dimension. Building on this observation, we propose \textbf{Flash EQ-Linear}, an exact acceleration algorithm that reduces the cost to $2(T-1)/T^2$ of the original dense formulation ($T$ is the equivariant group size) by combining the Fourier convolution theorem along the group dimension with the conjugate symmetry of the real DFT. To translate these computational savings into wall-clock speedups, we further develop dedicated CUDA kernels  for the $\mathrm{p}4$ group. At the operator level, Flash EQ-Linear achieves up to $\bm{2.1\times}$ forward speedup over PyTorch's highly optimized \texttt{F.linear}; at the network level, Flash EQ-ViT achieves up to $\bm{1.7\times}$ end-to-end speedup over both equivariant and non-equivariant baselines. As an operator-level acceleration algorithm, Flash EQ-Linear provides plug-and-play acceleration for diverse pretrained equivariant models, including EQ-ViT, EQ-Swin, EQ-VMamba, and EQ-INR, without retraining or architectural changes. Code is available at \url{https://github.com/zhongchenzhao/FlashEQLinear}.
\end{abstract}

\section{Introduction}

Equivariant networks provide a principled way to explicitly embed geometric symmetry priors (e.g., translation, rotation, and reflection) into network architectures by sharing weights across a transformation group~\citep{ravanbakhsh2017equivariance}.
Rather than learning such symmetries implicitly from augmented data, they guarantee by construction that a transformation of the input induces the corresponding transformation of the output, which improves robustness and generalization while using substantially fewer parameters.
Equivariant variants now span the major vision architecture families: CNNs~\citep{cohen2016group, cohen2016steerable, weiler2018learning, kondor2018generalization, weiler2019general, shen2020pdoeconvs, shen2021pdo, he2021efficient, xie2022fourier}, Vision Transformers (EQ-ViTs)~\citep{romero2021group, xu2023e2vit, fu2026vanilla}, Visual Mamba (EQ-VMamba)~\citep{zhao2026rotation}, and implicit neural representations (EQ-INR)~\citep{xie2025rotation}.
Each serves as a plug-and-play replacement for its standard counterpart, matching or exceeding its accuracy with far fewer parameters across image classification, segmentation, and restoration.

\begin{figure}[t]
	\centering
	\includegraphics[width=0.999\linewidth]{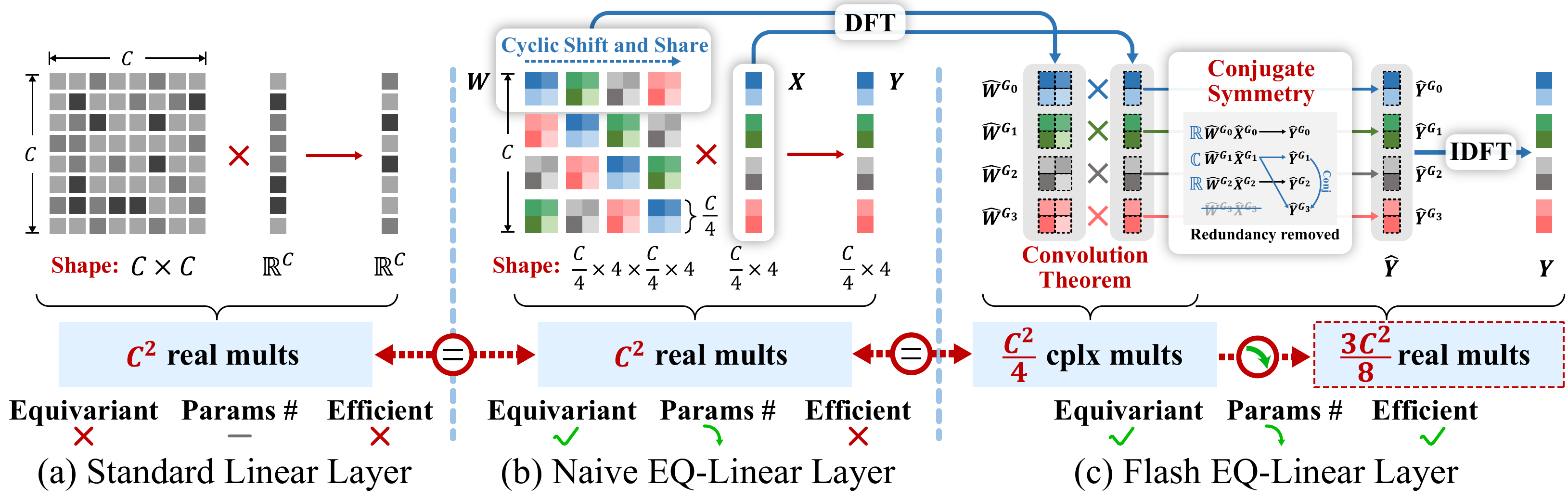}
	\vspace{-5.5mm}
	\caption{Illustration of the EQ-Linear under the $\mathrm{p}4$ group. The weight matrix ${\bm W}$ is constructed by cyclically shifting and replicating a set of learnable parameters $\tilde{\bm W}$ along the group dimension.}
	\label{fig:eq_linear}
	\vspace{-5.5mm}
\end{figure}

Despite the remarkable parameter efficiency and performance, equivariant networks generally suffer from a critical bottleneck: \textbf{the parameter efficiency has not translated well into compute efficiency}. 
The cause lies in how weight sharing is realized: the shared weights are replicated, permuted, and tiled along the group dimension into a dense weight tensor, which is then handed to a generic dense kernel.
This naive replication-and-dense-multiplication strategy incurs substantial execution overhead, making equivariant networks \emph{slower} than their non-equivariant counterparts~\citep{gerken2022equivariance}.
While a few recent efforts~\citep{nordstrom2026octic} have designed customized efficient ViT architectures, their structural modifications cannot be easily transferred to more general equivariant network families, nor can they be directly applied to pre-trained equivariant networks for acceleration, which largely limits their generality. 
Moreover, current equivariant layers lack the highly optimized CUDA kernels~\citep{chetlur2014cudnn} that standard network layers benefit from, which makes them hard to deploy in practice.

Ideally, an acceleration technique for equivariant models should target the operator level, specifically the fundamental equivariant linear (EQ-Linear) layer~\citep{finzi2021practical, xie2025rotation}. 
EQ-Linear is a core building block of EQ-ViTs, EQ-VMamba, and EQ-INR, and dominates their computational cost.
In EQ-ViT, for instance, EQ-Linear layers appear in every attention projection and every FFN block, accounting for more than 50\% of both FLOPs and inference latency (Table~\ref{tab:flash_EQViT_inference}). 
EQ-Linear is thus the main computational bottleneck shared by these backbones, and accelerating it at the operator level benefits all of them at once, without redesigning or retraining any of them.

However, the current naive implementation of EQ-Linear suffers from dense matrix-multiply and repetitive memory movement (as shown in Fig.~\ref{fig:eq_linear}(b)). In particular, the underlying parameter sharing mechanism not only fails to deliver computational efficiency, but severely degrades the execution speed due to repeated memory access overhead. 
Although current workarounds typically pre-assemble the weight matrix at inference, allowing EQ-Linear to run exactly as fast as a vanilla linear layer, this merely hides the redundancy rather than translating parameter efficiency into compute efficiency.

Instead of treating the EQ-Linear weight matrix as an ordinary dense matrix, we observe that it exhibits a \textbf{Cyclic Shift and Share} structure: the shared parameters are cyclically shifted and replicated $T$ times along the group axis.
This structural regularity underlies our key insight: EQ-Linear performs a circular convolution along the group dimension composed with a linear transformation along the channel dimension, and a circular convolution costs far less than a dense matrix product.

This observation unlocks a significantly faster mathematical formulation for EQ-Linear.
\textbf{(i)} By the convolution theorem of the discrete Fourier transform (DFT), the circulant convolution along the group dimension is exactly a pointwise multiplication in the Fourier domain~\citep{oppenheim2009discrete}, replacing $NC^{2}$ real multiplications with $\nicefrac{NC^{2}}{T}$ complex ones, where $N$ is the token count, $C$ the channel width, and $T$ the group size.
\textbf{(ii)} By the conjugate symmetry of the real DFT, about half of the frequency components are conjugates of the other half and need not be computed, while the zero and Nyquist ($\pi$) frequency components are purely real and require only real arithmetic.
Together these reduce the computation of EQ-Linear to approximately $\nicefrac{2(T-1)NC^{2}}{T^{2}}$ real multiplications.

Building on the above analysis, we propose \textbf{Flash EQ-Linear}, an exact acceleration algorithm that delivers an approximately $\boldsymbol{2\times}$ speedup for $\mathrm{p}4$ equivariant linear layers in the compute-bound regime. 
Unlike prior efforts that redesign equivariant architectures, Flash EQ-Linear accelerates the existing EQ-Linear operator itself with dedicated CUDA kernels, so it applies to pretrained p4-equivariant models without retraining.
The main contributions of this paper can be summarized as follows:

\begin{itemize}[leftmargin=*, itemsep=2pt, topsep=2pt]
	\item \textbf{Algorithmic.}
	We propose {Flash EQ-Linear}: an exact acceleration algorithm for the equivariant linear layer.
	It reformulates group convolution as pointwise multiplication in the frequency domain and further exploits the conjugate symmetry of the real DFT to remove redundant frequency-domain operations, reducing the theoretical cost to $\nicefrac{2(T-1)}{T^{2}}$ of the original dense formulation\footnote{The stated ratio assumes an even group size $T$, as is standard in practice; for odd group size, it is $\nicefrac{(2T-1)}{T^{2}}$.}.
	
	\item \textbf{Exactness and transferability.}
	The proposed algorithm is mathematically exact: relying only on the invertibility of the DFT and the convolution theorem, its outputs are identical to those of the naive EQ-Linear. 
	Besides, it serves as a plug-and-play replacement for the EQ-Linear layers of pretrained equivariant networks across architecture families, including Transformers (EQ-ViT and EQ-Swin), state space models (EQ-VMamba), and implicit neural representations (EQ-INR), requiring no fine-tuning or architectural change.

	\item \textbf{Hardware-aware CUDA kernels.}
	We develop dedicated CUDA kernels for Flash EQ-Linear that support both forward and backward passes in FP32 and FP16, translating theoretical computational savings into practical speedups.
	The kernels fuse the group-wise transforms into the prologue and epilogue of a GEMM-like channel reduction, keeping every frequency-domain intermediate in registers and shared memory, and are released as an out-of-the-box library for the community.
	
	\item \textbf{Empirical validation.}
	At the operator level, Flash EQ-Linear achieves up to $\boldsymbol{2.1\times}$ forward speedup over PyTorch's \texttt{F.linear} in the compute-bound regime.
	At the network level, Flash EQ-ViT achieves up to $\boldsymbol{1.7\times}$ end-to-end speedup over standard ViT and the corresponding equivariant baselines.
	These results follow the trend predicted by our theoretical analysis.
\end{itemize}

\section{Related Work}

\textbf{Group equivariant networks.}
Group equivariance was introduced to deep learning by G-CNN~\citep{cohen2016group}, which generalizes convolution by sharing weights across a transformation group.
Subsequent work broadened the supported groups~\citep{cohen2016steerable, weiler2019general}, refined the parameterization of group representations~\citep{xie2022fourier, xie2025rotation}, and carried equivariance into modern architecture families, including ViTs~\citep{romero2021group, xu2023e2vit, fu2026vanilla}, VMamba~\citep{zhao2026rotation}, and implicit neural representations~\citep{xie2025rotation}. 
Most existing research, however, focuses on expressivity, generalization, and parameter efficiency, while overlooking the run-time efficiency of equivariant operators.

\textbf{Exact acceleration.}
Unlike pruning, quantization, or distillation~\citep{cheng2017survey, han2015deep, hinton2015distilling}, which trade accuracy for speed, exact acceleration leaves the computed function unchanged, achieving its gains through algebraic reformulation or hardware-aware optimization.
For algebraic reformulation, \citet{mathieu2013fast} computed CNN convolutions as pointwise products via the FFT~\citep{cooley1965algorithm}, with optimized CUDA implementations by \citet{vasilache2014fast}.
For hardware-aware optimization, FlashAttention~\citep{dao2022flashattention} accelerates exact attention through IO-aware tiling to reduce data movement across the GPU memory hierarchy.
Flash EQ-Linear combines both approaches: it applies the Fourier transform along the \emph{group} dimension rather than the \emph{spatial} dimension, and realizes the resulting reduction in fused CUDA kernels.

\textbf{Acceleration of equivariant networks.}
Most closely related, \citet{bokman2025flopping} and \citet{nordstrom2026octic} proposed customized equivariant ViT variants that achieve computational reductions by the isotypic decomposition. 
While this represents an important step, such solutions are architecture-specific and lack comprehensive system-level CUDA optimizations. 
More importantly, because they modify the underlying architecture, these methods invalidate existing model weights and require costly retraining from scratch.
In contrast, Flash EQ-Linear provides a plug-and-play operator-level acceleration for pretrained equivariant networks, without architectural changes or retraining.

\section{Flash EQ-Linear: Principle, Algorithm, and Implementations}

In this section, we develop Flash EQ-Linear, an exact algorithm that exploits the discrete Fourier transform (DFT) to reduce the complexity of  EQ-Linear. 
We first present the underlying acceleration principle based on the convolution theorem and the conjugate symmetry of the real DFT (Sec.~\ref{subsec:Acceleration_Principle}), then derive the algorithm and analyze its properties (Sec.~\ref{subsec:Algorithm_Complexity}).
Finally, we describe the mathematical specialization and CUDA kernel implementation for the $\mathrm{p}4$ group (Sec.~\ref{subsec:p4_implementation}).

\textbf{Formulation setup.} For clarity, we formulate our methodology under four simplifying conditions: (1) equal input and output channel widths, denoted by $C$; (2) an even group size $T$; (3) $C$ divisible by $T$, with $C \gg T$; and (4) a total token count $N \gg T$.
These assumptions reflect common settings and can be generalized to broader scenarios, such as asymmetric channel dimensions and odd $T$, essentially without altering our theoretical speedup derivations.
Furthermore, since practical group sizes are typically small (e.g., $T \in \{2, 4, 8\}$), any indivisibility can be easily resolved via negligible channel padding with minimal computational overhead.

\subsection{Acceleration Principle}
\label{subsec:Acceleration_Principle}

\textbf{Formulation of EQ-Linear.}
As illustrated in Fig.~\ref{fig:eq_linear}, the weight matrix ${\bm W}$ of an equivariant linear layer has a group-circulant structure formed by cyclically shifting a set of learnable parameters $\tilde{\bm W} \!=\! [\tilde{\bm W}_0, \tilde{\bm W}_{1}, \ldots, \tilde{\bm W}_{T-1}]$ along the group dimension.
Given a group-structured input ${\bm X} \! \in \! \mathbb{R}^{N\times \frac{C}{T}  \times  T}$ over a group $\mathcal{G} \!=\! \{G_t \! \mid \! t \!=\! 0,1,\ldots,T \! -\!1\}$, EQ-Linear maps it to an output  ${\bm Y} \! \in \! \mathbb{R}^{N\times\frac{C}{T} \times T}$ as
\begin{equation}\label{eq:EQLinear}
	\left[ \!\!
	\begin{array}{c}
		\bm Y_n^{G_0} \\
		\bm Y_n^{G_1} \\
		\vdots \\
		\bm Y_n^{G_{T-1}}
	\end{array}
	\!\! \right]
	\!\!  = \!\! 
	\left[ \!\!
	\begin{array}{cccc} 
		\tilde{\bm W}_0 & \tilde{\bm W}_{T-1} & \cdots & \tilde{\bm W}_{1} \\
		\tilde{\bm W}_{1} & \tilde{\bm W}_0 & \cdots & \tilde{\bm W}_{2} \\
		\vdots & \vdots & \ddots & \vdots \\
		\tilde{\bm W}_{T-1} & \tilde{\bm W}_{T-2} & \cdots & \tilde{\bm W}_0
	\end{array}
	\!\! \right] \! \! 
	\left[ \!\!
	\begin{array}{c}
		\bm X_n^{G_0} \\
		\bm X_n^{G_1} \\
		\vdots \\
		\bm X_n^{G_{T-1}}
	\end{array}
	\!\! \right]
	\! + \!
	\left[ \!\!
	\begin{array}{c}
		\tilde{\bm b} \\
		\tilde{\bm b} \\
		\vdots \\
		\tilde{\bm b}
	\end{array}
	\!\! \right] \!\!, 
	~~\mbox{for}~ n = 0, \ldots, N-1.
\end{equation}
Here, ${\bm X}_n^{G_t}, {\bm Y}_n^{G_t} \!\in\! \mathbb{R}^{\frac{C}{T}}$ denote the $t$-th group components\footnote{$G_t \!\in\! \mathcal{G}$ denotes both a group transformation (e.g., a rotation matrix) and an index along the group dimension. } of ${\bm X}_n, {\bm Y}_n \!\in\! \mathbb{R}^{\frac{C}{T}\times T}$, respectively.
The weight matrix ${\bm W} \!\in\! \mathbb{R}^{C \! \times \! C}$ and bias ${\bm b} \!\in\! \mathbb{R}^{C}$ are tiled from the learnable parameters $\tilde{\bm W} \!\in\! \mathbb{R}^{\frac{C}{T} \times C}$ and $\tilde{\bm b} \!\in\! \mathbb{R}^{\frac{C}{T}}$. 
As formulated in Eq.~\ref{eq:EQLinear}, the naive implementation ignores the group-circulant structure and expands $\tilde{\bm W}$ into a dense matrix, requiring $NC^2$ multiply-accumulate operations (MACs).

\textbf{EQ-Linear as a group-circulant convolution.}
Our key observation is that the EQ-Linear in Eq.~\ref{eq:EQLinear} essentially performs a circular convolution between each input token $\bm{X}_n$ and the parameters $\tilde{\bm{W}}$ along the group dimension, composed with matrix multiplications along the channel dimension:
\begin{equation}\label{eq:eqlinear_conv_form}
	\bm{Y}_n = \tilde{\bm{W}} \circledast_{\mathcal{G}} \bm{X}_n + \tilde{\bm{b}}, \quad \text{where} ~~ (\tilde{\bm{W}} \circledast_{\mathcal{G}} \bm{X}_n)^{G_t} = \sum_{s=0}^{T-1}   \tilde{\bm{W}}_{(t-s)\,\mathrm{mod}\,T} \cdot \bm{X}_n^{G_s}, ~~ \forall G_t \in \mathcal{G}.
\end{equation}
\textbf{Acceleration principle.}
This group-circulant structure unlocks a significantly faster mathematical formulation, leveraging two key properties of the DFT (as illustrated in Fig.~\ref{fig:principle_diagram}):

\begin{figure}[t]
	\centering
	\includegraphics[width=0.98	\linewidth]{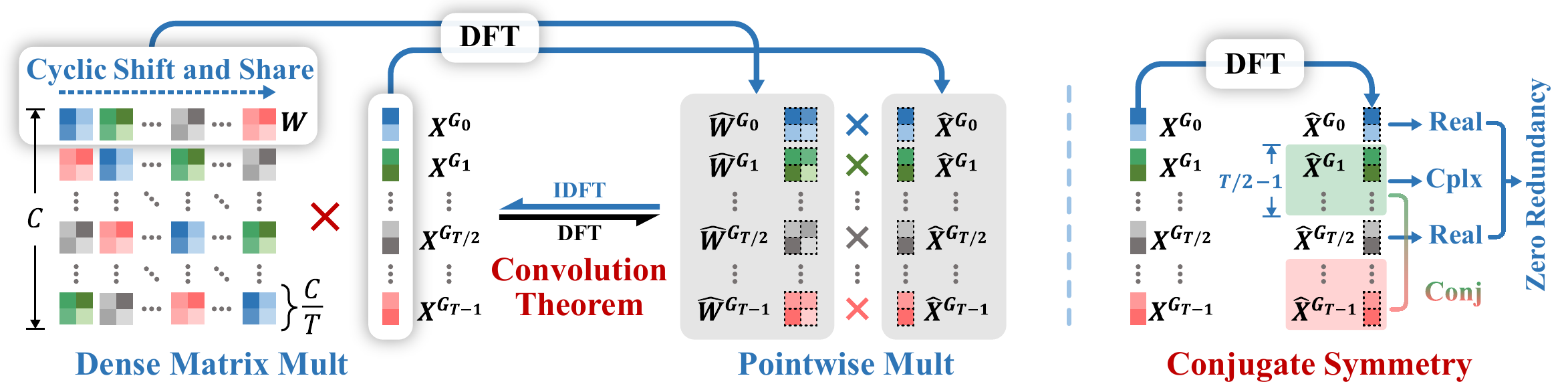}  
	\vspace{-2mm}
	\caption{ Acceleration  via the real DFT. \textbf{Left:} The convolution theorem maps dense EQ-Linear computation to pointwise multiplications across $T$ frequency bins. \textbf{Right:} Conjugate symmetry yields only $ T/2 +1$ non-redundant bins: frequency components at $k=0$ and $k=T/2$ are purely real, while the remaining components form conjugate pairs.
	}
	\label{fig:principle_diagram}
	\vspace{-5mm}
\end{figure}

\textbf{(i) Convolution Theorem:}
The classical convolution theorem states that a circular convolution corresponds to pointwise scalar multiplication in the frequency domain. In our case, Eq.~\ref{eq:eqlinear_conv_form} defines a mixed operation: circular convolution along the group dimension and matrix multiplication along the channel dimension. Since these two linear operations act on orthogonal dimensions, they commute. Consequently, applying the group-wise DFT pushes through the channel-wise matrix multiplications. As illustrated in the Left part of Fig.~\ref{fig:principle_diagram}, this strictly converts the dense EQ-Linear computation into $T$ independent per-frequency matrix multiplications:
\begin{equation}\label{eq:eqlinear_DFT}
	\mathcal{F}_{\mathcal{G}} \left(  \tilde{\bm W} \circledast_{\mathcal{G}} {\bm X}_n  \right)^{G_k} = \hat{\bm W}^{G_k} \cdot \hat{\bm X}_n^{G_k}, \quad \text{where} ~~ \hat{\bm W} = \mathcal{F}_{\mathcal{G}}(\tilde{\bm W}), \;\; \hat{\bm X}_n = \mathcal{F}_{\mathcal{G}}({\bm X}_n),
\end{equation}
for all $k \in \{0, \ldots, T-1\}$. Here, $\mathcal{F}_{\mathcal{G}}$ is formally defined as the 1D DFT applied strictly along the group dimension (e.g., ${\mathcal{F}_{\mathcal{G}}(\bm X_n)}^{G_k} = \sum_{t=0}^{T-1} \bm{X}_n^{G_t} e^{-\mathrm{i} 2\pi k t / T}$). Each independent frequency-domain multiplication maps $\frac{C}{T}$ input channels to $\frac{C}{T}$ output channels.

\textbf{(ii) Conjugate symmetry of real DFT:}
Since both $\bm{X}$ and $\tilde{\bm{W}}$ are real-valued, their Fourier coefficients exhibit conjugate symmetry:
\begin{equation}
	\label{eq:conj_sym}
	\hat{\bm X_n}^{G_{T-k}} = \overline{\hat{\bm X_n}^{G_k}}, \quad \forall k \in \{1, \ldots, T-1\},
\end{equation}
and likewise for $\hat{\bm W}$.
As illustrated in the Right part of Fig.~\ref{fig:principle_diagram}, nearly half of the frequency components are redundant. Given our setup of an even group size $T$, only the first $T/2 + 1$ non-redundant components ($k=0, \ldots, T/2$) need to be explicitly computed to achieve zero redundancy.
More importantly, the zero frequency ($k=0$) and the Nyquist frequency ($k=T/2$) components are purely real.
This allows us to substitute complex matrix multiplications ($4$ real-valued MACs) with standard real matrix multiplications ($1$ real-valued MAC) at these frequencies.

\subsection{Algorithm and Complexity Analysis}
\label{subsec:Algorithm_Complexity}

\begin{figure}[t]
	\centering\vspace{-2mm}
	\includegraphics[width=1.0\linewidth]{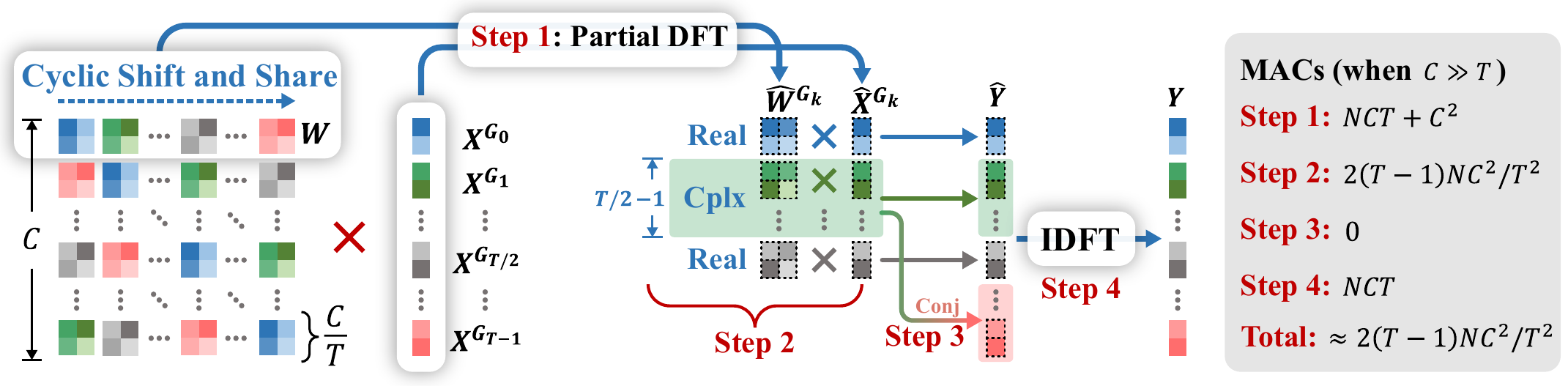}  
	\vspace{-6mm}
	\caption{\textbf{Execution flow of Flash EQ-Linear.} The computation proceeds in four stages: (\textbf{Step 1}) partial group-wise DFT; (\textbf{Step 2}) independent frequency-domain matrix multiplications, using real arithmetic for zero/Nyquist frequencies and complex arithmetic for remaining pairs; (\textbf{Step 3}) zero-cost conjugate-symmetric recovery; and (\textbf{Step 4}) group-wise IDFT with bias addition.}\vspace{-3mm}
	\label{fig:main_algorithm_diagram}
	\vspace{-2mm}
\end{figure}

\begin{algorithm}[t]
	\caption{Flash EQ-Linear (For even group size $T$)}
	\label{alg:flash_eqlinear}
	\renewcommand{\algorithmicrequire}{\textbf{Input:}}
	\renewcommand{\algorithmicensure}{\textbf{Output:}}
	\begin{spacing}{1.1}
		\small
		\begin{algorithmic}[1]
			\Require Real-valued input tokens ${\bm X} \in \mathbb{R}^{N \times \frac{C}{T} \times T}$, parameters $\tilde{\bm W} \in \mathbb{R}^{\frac{C}{T} \times \frac{C}{T} \times T}$, and bias $\tilde{\bm b} \in \mathbb{R}^{\frac{C}{T}}$.
			\State \textbf{Step 1:  Partial Group-wise DFT.} Compute DFT for non-redundant frequencies: 
			\Statex $\hat{\bm X}_n^{G_k} = \mathcal{F}_{\mathcal{G}}({\bm X_n})^{G_k}, \quad
			\hat{\bm W}^{G_k} = \mathcal{F}_{\mathcal{G}}(\tilde{\bm W})^{G_k},  ~~\mbox{for}~~ k=0,\ldots, T/2 ~~\mbox{and}~~  n = 0, \ldots, N-1.$
			\State \textbf{Step 2: Parallelizable per-frequency matrix multiplication.}
			\Statex \quad\quad\quad \textbf{(a) Pure real frequencies} ($2$ real matrix multiplications): 
			\Statex \quad\quad\quad \quad $\hat{\bm Y}_n^{G_{k}} = {{\hat{\bm W}}^{G_{k}}} \cdot \hat{\bm X_n}^{G_{k}},  ~~\mbox{for}~~ k=0, T/2 ~~\mbox{and}~~  n = 0, \ldots, N-1.$ 
			\Statex \quad\quad\quad \textbf{(b) Complex frequencies} ($T/2 - 1$ complex matrix multiplications):
			\Statex \quad\quad\quad \quad $\hat{\bm Y}_n^{G_{k}} = {{\hat{\bm W}}^{G_{k}}} \cdot \hat{\bm X_n}^{G_{k}}, ~~\mbox{for}~~ k=1, \ldots, T/2-1 ~~\mbox{and}~~  n = 0, \ldots, N-1.$ 
			\State \textbf{Step 3: Conjugate-symmetric recovery.} Recover $k = T/2 + 1, \ldots, T-1$ via $\hat{\bm Y}^{G_{k}} = \overline{\hat{\bm Y}^{G_{T-k}}}$.
			\State \textbf{Step 4: Group-wise IDFT.} Apply inverse DFT to $\hat{\bm Y}$ and add bias $\tilde{\bm b}$.
			\Ensure ${\bm Y} \in \mathbb{R}^{N \times \frac{C}{T} \times T}$
		\end{algorithmic}
	
	\end{spacing}
\end{algorithm}
Building on the dual acceleration principle, we formally propose the \textbf{Flash EQ-Linear algorithm}. Specifically, the algorithm leverages the convolution theorem (Eq.~\ref{eq:eqlinear_DFT}) to decompose the dense matrix multiplication into a series of compact, independent frequency-domain matrix multiplications, and explicitly exploits conjugate symmetry (Eq.~\ref{eq:conj_sym}) to further eliminate redundant computations. 
The complete execution pipeline is detailed in Algorithm~\ref{alg:flash_eqlinear} and illustrated in Fig.~\ref{fig:main_algorithm_diagram}, which map the theoretical derivations into an efficient four-step computational flow.

\textbf{Complexity analysis.}
Among the four steps of Algorithm~\ref{alg:flash_eqlinear}, the per-frequency matrix multiplications (Step 2) constitute the primary computational bottleneck.
Flash EQ-Linear reduces this dominant cost by exploiting the conjugate symmetry made explicit in Fig.~\ref{fig:main_algorithm_diagram}: two self-conjugate frequencies $k=0$ and $k=T/2$ are purely real and reduce to real matrix multiplications, whereas each of the remaining $T/2-1$ frequencies is complex and requires four real matrix multiplications.
Step~2 therefore costs $2 \frac{NC^2}{T^2} + 4(T/2-1)\frac{NC^2}{T^2} =\frac{2(T-1)NC^2}{T^2}$ real-valued MACs.
We formalize the resulting efficiency gain in Theorem~\ref{theorem:MACs_Flash_EQ-Linear}, with the full derivation deferred to Appendix~\ref{app:appendix_more_Complexity_analysis}.
\begin{Theorem}[Complexity of Flash EQ-Linear]
	\label{theorem:MACs_Flash_EQ-Linear}
	Let ${\bm X}, {\bm Y} \in \mathbb{R}^{N \times \frac{C}{T} \times T}$ be the input and output of an EQ-Linear layer with an even group size $T$. When $C \gg T$ and $N \gg T$, the number of real-valued MACs incurred by Flash EQ-Linear satisfies
	\begin{equation}
		\mathrm{MACs}_{\mathrm{Flash}} \approx 2(T-1) N C^2/{T^2}.
	\end{equation}
	Compared with the $NC^2$ MACs of naive EQ-Linear, this yields a theoretical speedup of $\nicefrac{T^2}{2(T-1)}$.
\end{Theorem}
For the most ubiquitous transformation groups in practice, such as $90^\circ$ rotations ($T=4$) and mirror reflections ($T=2$), this translates to concrete speedups of $\approx \! 2.67\times$ and $2\times$, respectively.

\textbf{Properties.} Beyond the asymptotic reduction in computational complexity, the proposed Flash EQ-Linear design inherently enjoys several highly desirable properties for practical deployment: 

\textbf{(i)  Exactness and equivariance preservation.}
Flash EQ-Linear is exact by construction: the invertibility of DFT and the convolution theorem make it algebraically equivalent to naive EQ-Linear.
It therefore preserves the rotation equivariance, as verified in Tables~\ref{tab:flash_eqlinear_Precision_Errors} and \ref{tab:eq_error} in the Appendix.

\textbf{(ii) Speedup scales with group size.}
As established in Theorem~\ref{theorem:MACs_Flash_EQ-Linear}, the theoretical speedup $\nicefrac{T^2}{2(T-1)}$ of Flash EQ-Linear grows with the group size $T$, from $2\times$ for $T \! = \! 2$ to  $4.57\times$ for $T \! = \! 8$.
Thus, EQ-Linear with finer groups (larger $T$) admit greater speedups, as verified in Table~\ref{tab:flash-eqlinear-T-scaling} in the Appendix.

\textbf{(iii) Training-free and plug-and-play.}
Flash EQ-Linear can be used to directly replace naive EQ-Linear in pretrained equivariant networks in a plug-and-play manner, requiring no retraining, fine-tuning, or architectural changes.
Existing models such as EQ-ViT and EQ-Swin can therefore adopt Flash EQ-Linear without modifying their training or inference pipelines (as shown in Table~\ref{tab:flash_EQSwin_inference}).

\subsection{Implementation for the $\mathrm{p}4$ Group}
\label{subsec:p4_implementation}

For the widely used $\mathrm{p}4$ rotation group~\citep{cohen2016group} ($T=4$), specializing Alg.~\ref{alg:flash_eqlinear} yields a substantially simplified form that maps cleanly onto an efficient fused implementation.

\textbf{Step 1: Group-wise DFT.}
For the $\mathrm{p}4$ rotation group ($T=4$), the group-wise discrete Fourier transform of $\bm X$ is specialized to the following explicit form:
\begin{equation}\label{eq:DFT_p4}
	\small
	\setlength{\arraycolsep}{4pt}
	\begingroup
	\renewcommand{\arraystretch}{1.25}
	\begin{bmatrix}
		\hat{\bm{X}}_n^{G_0} \\
		\hat{\bm{X}}_n^{G_1} \\
		\hat{\bm{X}}_n^{G_2} \\
		\hat{\bm{X}}_n^{G_3}
	\end{bmatrix}
	\endgroup
	\!\! = \!\!
	\begingroup
	\renewcommand{\arraystretch}{1.25}
	\begin{bmatrix}
		1 & 1  & 1  & 1  \\
		1 & -i & -1 & i  \\
		1 & -1 & 1  & -1 \\
		1 & i  & -1 & -i
	\end{bmatrix}
	\endgroup
	\!\!
	\begingroup
	\renewcommand{\arraystretch}{1.25}
	\begin{bmatrix}
		\bm{X}_n^{G_0} \\
		\bm{X}_n^{G_1} \\
		\bm{X}_n^{G_2} \\
		\bm{X}_n^{G_3}
	\end{bmatrix}
	\endgroup
	\! \implies \!
	\begin{bmatrix}
		\hat{\bm{X}}_n^{G_0} \\
		\operatorname{Re}(\hat{\bm{X}}_n^{G_1}) \\
		\operatorname{Im}(\hat{\bm{X}}_n^{G_1})\\
		\hat{\bm{X}}_n^{G_2} \\
	\end{bmatrix}
	\!\! = \!\!
	\begingroup
	\renewcommand{\arraystretch}{1.1}
	\begin{bmatrix}
		1 & 1  & 1  & 1  \\
		1 & 0  & -1 & 0  \\
		0 & -1 & 0  & 1\\
		1 & -1 & 1  & -1 
	\end{bmatrix}
	\endgroup
	\!\!
	\begin{bmatrix}
		\bm{X}_n^{G_0} \\
		\bm{X}_n^{G_1} \\
		\bm{X}_n^{G_2} \\
		\bm{X}_n^{G_3}
	\end{bmatrix}
	\, .
\end{equation}
The reduced DFT transform matrix in Eq.~\ref{eq:DFT_p4} contains only $0$ and $\pm1$.
Consequently, the group-wise DFT reduces entirely to additions and subtractions, requiring no nontrivial multiplications.

\textbf{Step 2: Per-frequency Fourier-domain matrix multiplication.}
Using the four real-valued Fourier components derived in Step~1, the per-frequency multiplication specializes to
\begin{equation}\label{eq:appendix_p4_step2}
	\begin{aligned}
		&\hat{\bm Y}_n^{G_0} =  \hat{\bm W}^{G_0} \cdot \hat{\bm X}_n^{G_0}, ~~ 
		\hat{\bm Y}_n^{G_2} = \hat{\bm W}^{G_2} \cdot \hat{\bm X}_n^{G_2}, ~~ \hat{\bm Y}_n^{G_3} =\overline{\hat{\bm Y}_n^{G_1}}, \\
		&\mathrm{Re}(\hat{\bm Y}_n^{G_1}) = \mathrm{Re}(\hat{\bm W}^{G_1}) \cdot \mathrm{Re}(\hat{\bm X}_n^{G_1})-\mathrm{Im}(\hat{\bm W}^{G_1}) \cdot \mathrm{Im}(\hat{\bm X}_n^{G_1}), \\
		&\mathrm{Im}(\hat{\bm Y}_n^{G_1}) = \mathrm{Im}(\hat{\bm W}^{G_1}) \cdot \mathrm{Re}(\hat{\bm X}_n^{G_1}) + \mathrm{Re}(\hat{\bm W}^{G_1}) \cdot \mathrm{Im}(\hat{\bm X}_n^{G_1}).
	\end{aligned}
\end{equation}
Step~2 thus reduces to six real-valued matrix multiplications, which can be executed in parallel.
Since each multiplication maps ${\frac{C}{T}}$ input channels to ${\frac{C}{T}}$ output channels, the total cost is $\bm{\frac{6}{16}NC^2}$ MACs.

\textbf{Step 3: Conjugate-symmetric recovery.}
Real-valued inputs and weights imply $\hat{\bm Y}^{G_3}=\overline{\hat{\bm Y}^{G_1}}$, allowing the inverse transform to incorporate $G_3$ implicitly without materializing it.

\textbf{Step 4: Group-wise IDFT.}
Exploiting the conjugate symmetry from Step~3, the group-wise inverse DFT for the $\mathrm{p}4$ rotation group can be expressed directly using four real-valued Fourier components:
\begin{equation}\label{eq:appendix_IDFT}
	\small
	\setlength{\arraycolsep}{4pt}
	\begin{bmatrix}
		{\bm{Y}}_n^{G_0} \\
		{\bm{Y}}_n^{G_1} \\
		{\bm{Y}}_n^{G_2} \\
		{\bm{Y}}_n^{G_{3}}
	\end{bmatrix}
	= 
	\frac{1}{4}
	\begingroup
	\renewcommand{\arraystretch}{1.25}
	\begin{bmatrix}
		1 & 1 & 1 & 1 \\
		1 & i & -1 & -i \\
		1 & -1 & 1 & -1 \\
		1 & -i & -1 & i
	\end{bmatrix}
	\endgroup 
	\begingroup
	\renewcommand{\arraystretch}{1.25}
	\begin{bmatrix}
		{\hat{\bm{Y}}}_n^{G_0} \\
		{\hat{\bm{Y}}}_n^{G_1} \\
		{\hat{\bm{Y}}}_n^{G_2} \\
		{\hat{\bm{Y}}}_n^{G_{3}}
	\end{bmatrix}
	\endgroup
	=
	\frac{1}{4}
	\begingroup
	\renewcommand{\arraystretch}{1.25}
	\begin{bmatrix}
		1 & 2 & 0 & 1 \\
		1 & 0 & -2 & -1 \\
		1 & -2 & 0 & 1 \\
		1 & 0 & 2 & -1
	\end{bmatrix}
	\endgroup
	\begingroup
	\renewcommand{\arraystretch}{1.25}
	\begin{bmatrix}
		{\hat{\bm{Y}}}_n^{G_0} \\
		\mathrm{Re}(\hat{\bm{Y}}_n^{G_1}) \\
		\mathrm{Im}(\hat{\bm{Y}}_n^{G_1}) \\
		{\hat{\bm{Y}}}_n^{G_2}
	\end{bmatrix}
	\endgroup\!.
\end{equation}
Eq.~\ref{eq:appendix_IDFT}  contains only the dyadic coefficients $\{0,\pm\frac{1}{4},\pm\frac{1}{2}\}$ and thus reduces to additions, subtractions, and power-of-two rescalings, incurring no additional MACs in the final implementation.

\textbf{CUDA kernel design.}
A straightforward PyTorch implementation of Alg.~\ref{alg:flash_eqlinear} decomposes the computation into separate DFT, complex multiplication, and IDFT operators.
The resulting kernel launches, layout transformations, and global-memory traffic can dominate runtime and diminish the algorithmic gains~\citep{williams2009roofline,dao2022flashattention}.
To translate the theoretical speedups into practical speedups, we develop dedicated CUDA kernels of Flash EQ-Linear for the $\mathrm{p}4$ rotation group, supporting both forward and backward passes in FP32 and FP16.

Our implementation fuses all four steps of Alg.~\ref{alg:flash_eqlinear} into a single GEMM-like dataflow, eliminating costly intermediate round trips to global memory.
For $\mathrm{p}4$ with $T=4$, the DFT and IDFT are fully unrolled into additions, subtractions, and fixed rescalings, with intermediate Fourier-domain fragments retained in registers or shared memory.
Weights are stored in frequency-domain planes, while input activations remain in their native layout, avoiding auxiliary packed tensors and additional GPU memory allocation.
We further provide shape-specialized kernel variants that tune tile sizes, shared-memory staging, persistence, Tensor Core usage, and vectorized memory access across different $(B, L, \frac{C}{T}, \frac{C}{T})$ regimes.
Further implementation details are provided in Appendix~\ref{sec:appendix_more_cuda_kernel}.

\section{Experiments}

In this section, we evaluate the practical speedups of Flash EQ-Linear through operator-level and network-level runtime benchmarks. 
Complete experimental results, including validation of numerical accuracy and equivariance preservation, are provided in Appendix~\ref{sec:appendix_more_results}.

\subsection{Operator-Level Acceleration}
\label{subsec:speed_eqlinear}

\textbf{Experimental settings.}
We benchmark Flash EQ-Linear against two baselines: standard Linear (PyTorch's \texttt{F.linear}~\citep{paszke2019pytorch}) and Naive EQ-Linear, which explicitly unrolls the equivariant weights. 
To quantify the benefits of the CUDA optimizations, we compare the PyTorch and CUDA implementations of Flash EQ-Linear.
All methods receive the same input feature map ${\bm X} \! \in \! \mathbb{R}^{B\times L\times \frac{C}{T} \times T}$, with batch size $B \!=\!32$, sequence length $L \!= \! 1024$, and group size $T \!= \!4$. 
We vary the per-group channel width $\nicefrac{C}{T}$ from $64$ to $2048$. 
For a fair comparison, the standard Linear applies a dense $C \! \to \! C$ matrix projection. 
Parameters and FLOPs are reported at $\nicefrac{C}{T}=64$. 
We measure forward and backward latency on a single NVIDIA RTX~4090 GPU, averaging over $200$ runs.

\begin{table*}[t]
	\centering
	\caption{
		\textbf{Forward Latency comparison} of Flash EQ-Linear against Standard Linear and Naive EQ-Linear baselines under different channel numbers ($\nicefrac{C}{T} \! = \! 64 \! \to \! 2048$).
		\textit{Speedup ratios} (\textcolor{green!50!black}{green}) are measured against the standard Linear (PyTorch's \texttt{F.linear}) baseline.
	}
	\vspace{-2mm}
	\label{tab:flash-eqlinear-forward}
	\newcommand{\speedup}[1]{\textcolor{green!50!black}{\textbf{#1$\times$}}}
	\newcommand{\smallspeedup}[1]{{\scriptsize\textcolor{green!50!black}{(\textbf{#1$\times$})}}}
	
	\setlength{\tabcolsep}{6.5pt}
	\small
	\begin{tabular}{lcc cccccc}
		\toprule
		\multirow{2.5}{*}{Method} 
		& \multirow{2.5}{*}{\makecell{\#Param.\\(M)$\downarrow$}} 
		& \multirow{2.5}{*}{\makecell{FLOPs\\(G)$\downarrow$}} 
		& \multicolumn{6}{c}{Forward Latency (ms)$\downarrow$} \\
		\cmidrule(lr){4-9} 
		& &  & 64 & 128 & 256 & 512 & 1024 & 2048 \\
		\midrule 
		\multicolumn{9}{c}{\textit{Single-Precision Floating-Point (FP32)}} \\
		\midrule 
		
		Standard Linear  & 0.066 & 0.134  & 0.11 & 0.43 & 1.52 & 6.05 & 24.48 & 104.96 \\
		Naive EQ-Linear & \textbf{0.016} & 0.134 & 0.11 & 0.44 & 1.54 & 6.25 & 25.16 & 105.74 \\
		Flash EQ-Linear (PyTorch) & \textbf{0.016} & \textbf{0.052} & 0.52 & 0.99 & 2.24 & 5.82 & 16.16 & 51.89 \\
		\rowcolor{gray!10}
		\textbf{Flash EQ-Linear (CUDA)}   
		& \textbf{0.016} & \textbf{0.052} 
		& \textbf{0.07} & \textbf{0.25} & \textbf{0.79} & \textbf{3.03} & \textbf{11.92} & \textbf{49.14} \\
		
		\rowcolor{gray!10}
		\textit{Speedup/Reduction $\uparrow$}  
		& \textbf{75.0\%} & \textbf{61.2\%} 
		& \speedup{1.6} & \speedup{1.7} & \speedup{1.9} & \speedup{2.0} & \speedup{2.1} & \speedup{2.1} \\
		
		\midrule 
		
		\multicolumn{9}{c}{\textit{Half-Precision Floating-Point (FP16)}} \\
		\midrule 
		Standard Linear            
		& 0.066 & 0.134          
		& 0.05 & 0.12 & 0.44 & 1.70 & 6.79 & 30.08 \\
		
		Naive EQ-Linear & \textbf{0.016} & 0.134  & 0.05 & 0.12 & 0.46 & 1.73 & 6.90 & 30.59 \\
		Flash EQ-Linear (PyTorch) & \textbf{0.016} & \textbf{0.052} & 0.58 & 0.61 & 1.20 & 2.68 & 6.73 & 18.67 \\
		
		\rowcolor{gray!10}
		\textbf{Flash EQ-Linear (CUDA)} & \textbf{0.016} & \textbf{0.052}  & \textbf{0.04} & \textbf{0.09} & \textbf{0.31} & \textbf{1.04} & \textbf{3.23} & \textbf{14.07} \\
		
		\rowcolor{gray!10}
		\textit{Speedup/Reduction $\uparrow$}  
		& \textbf{75.0\%} & \textbf{61.2\%}  
		& \speedup{1.3} & \speedup{1.3} & \speedup{1.4} & \speedup{1.6} & \speedup{2.1} & \speedup{2.1} \\
		\bottomrule
	\end{tabular}
\end{table*}

\begin{table*}[t]
	\centering
	\caption{
		\textbf{Backward latency comparison} of Flash EQ-Linear against Standard Linear and Naive EQ-Linear baselines under different channel numbers ($\nicefrac{C}{T} \! = \! 16 \! \to \! 2048$) under FP32.
	}
	\vspace{-2mm}
	\label{tab:flash_eqlinear_backward_fp32}
	\newcommand{\speedup}[1]{\textcolor{green!50!black}{\textbf{#1$\times$}}}
	\newcommand{\slowdown}[1]{\textcolor{gray!90!black}{\textbf{#1$\times$}}}
	\newcommand{\smallspeedup}[1]{{\scriptsize\textcolor{green!50!black}{(\textbf{#1$\times$})}}}
	\setlength{\tabcolsep}{7.5pt}
	\small
	\begin{tabular}{lcccccccc}
		\toprule
		\multirow{2.5}{*}{Method} 
		& \multicolumn{8}{c}{Backward Latency (ms)$\downarrow$} \\
		\cmidrule(lr){2-9} 
		& 16 & 32 & 64 & 128 & 256 & 512 & 1024 & 2048 \\
		\midrule 
		Standard Linear        & 0.17 & {0.14} & 0.16 & 0.46 & 1.67 & 5.42 & 20.80 & 83.11 \\
		Naive EQ-Linear        & 0.34 & 0.33 & 0.33 & 0.53 & 1.80 & 5.68 & 22.35 & 89.34 \\
		Flash EQ-Linear (PyTorch) & {2.88} & 3.18 & 4.17 & 8.43 & 15.63 & 30.62 & 67.49 & 170.92 \\
		\rowcolor{gray!10}
		\textbf{Flash EQ-Linear (CUDA)}  & \textbf{0.14} & \textbf{0.13} & \textbf{0.13} & \textbf{0.39} & \textbf{1.29} & \textbf{4.70} & \textbf{17.79} & \textbf{70.91} \\
		
		\rowcolor{gray!10}
		\textit{Speedup (vs. Standard) $\uparrow$}  
		& \speedup{1.2} & \speedup{1.1} & \speedup{1.3} & \speedup{1.2} & \speedup{1.3} & \speedup{1.2} & \speedup{1.2} & \speedup{1.2} \\
		
		\rowcolor{gray!10}
		\textit{Speedup (vs. Naive) $\uparrow$}  
		& \speedup{2.5} & \speedup{2.6} & \speedup{2.6} & \speedup{1.4} & \speedup{1.4} & \speedup{1.2} & \speedup{1.3} & \speedup{1.3} \\
		\bottomrule
	\end{tabular}
	\vspace{-4mm}
\end{table*}

\begin{table*}[t]
	\centering
	\caption{
		\textbf{Inference latency comparison} of Flash EQ-ViT against standard non-EQ and naive EQ baselines on ImageNet-100.
		\textit{Speedup ratios} (\textcolor{green!50!black}{green}) are measured against standard ViT baselines.
	}
	\vspace{-2mm}
	\label{tab:flash_EQViT_inference}
	\newcommand{\NA}{--}
	\newcommand{\flashrow}{\rowcolor{gray!10}}
	\setlength{\tabcolsep}{5.5pt}
	\renewcommand{\arraystretch}{1.12}
	\resizebox{\textwidth}{!}{
		\begin{tabular}{l c c l l l l l l}
			\toprule
			\multirow{3}{*}{\textbf{Method}}
			& \multirow{3}{*}{\makecell{\textbf{\#Param.}\\\textbf{(M)}$\downarrow$}}
			& \multirow{3}{*}{\makecell{\textbf{Top-1}\\\textbf{(\%)}$\uparrow$}}
			& \multicolumn{3}{c}{\textbf{All Linear Layers}}
			& \multicolumn{3}{c}{\textbf{Total Network}} \\
			\cmidrule(lr){4-6}
			\cmidrule(lr){7-9}
			&
			&
			& \makecell{\textbf{FLOPs}\\\textbf{(G)}$\downarrow$}
			& \makecell{\textbf{Latency}\\\textbf{(ms)}$\downarrow$}
			& \makecell{\textbf{Latency}\\\textbf{Ratio}}
			& \makecell{\textbf{FLOPs}\\\textbf{(G)}$\downarrow$}
			& \makecell{\textbf{Latency}\\\textbf{(ms)}$\downarrow$}
			& \makecell{\textbf{Throughput}\\\textbf{(imgs/s)}$\uparrow$} \\
			\midrule
			\multicolumn{9}{c}{\textit{Single-Precision Floating-Point (FP32)}} \\
			\midrule
			ViT-S                 & 33.8 & 76.5 & 13.1 & 0.352 & 58.4\% & 14.1 & 0.603  & 1658 \\
			Naive EQ-ViT-S        & \textbf{8.4}  & \textbf{78.2} & 13.1 & 0.355 & 58.7\% & 14.1 & 0.604 & 1656 \\
			\flashrow
			\textbf{Flash EQ-ViT-S}
			& \textbf{8.4} & \textbf{78.2} & \textbf{5.0} & \textbf{0.183} & \textbf{42.6\%}
			& \textbf{6.0} & \textbf{0.429} & \textbf{2332}\,{\scriptsize\textcolor{green!50!black}{(\textbf{1.4$\times$})}} \\
			ViT-B                 & 85.9 & 77.1 & 33.5 & 0.761 & 67.1\% & 35.1 & 1.135 & 881 \\
			Naive EQ-ViT-B        & \textbf{21.4} & \textbf{80.1} & 33.5 & 0.757 & 67.1\% & 35.1 & 1.128 & 886 \\
			\flashrow
			\textbf{Flash EQ-ViT-B}
			& \textbf{21.4} & \textbf{80.1} & \textbf{12.6} & \textbf{0.400} & \textbf{52.3\%}
			& \textbf{14.3} & \textbf{0.764} & \textbf{1309}\,{\scriptsize\textcolor{green!50!black}{(\textbf{1.5$\times$})}} \\
			ViT-L                 & 303.4 & 78.4 & 119.0 & 2.672 & 71.9\% & 123.1 & 3.715 & 269 \\
			Naive EQ-ViT-L        & \textbf{75.8} & \textbf{80.6} & 119.0 & 2.760 & 73.9\% & 123.1 & 3.735 & 268 \\
			\flashrow
			\textbf{Flash EQ-ViT-L}
			& \textbf{75.8} & \textbf{80.6} & \textbf{44.9} & \textbf{1.336} & \textbf{56.7\%}
			& \textbf{49.0} & \textbf{2.356} & \textbf{424}\,{\scriptsize\textcolor{green!50!black}{(\textbf{1.6$\times$})}} \\
			ViT-H                 & 631.0 & 81.3 & 248.0 & 5.586 & 77.7\% & 254.7 & 7.192 & 139 \\
			Naive EQ-ViT-H        & \textbf{157.7} & \textbf{81.5} & 248.0 & 5.627 & 78.4\% & 254.7 & 7.182 & 139 \\
			\flashrow
			\textbf{Flash EQ-ViT-H}
			& \textbf{157.7} & \textbf{81.5} & \textbf{93.4} & \textbf{2.719} & \textbf{63.2\%}
			& \textbf{100.1} & \textbf{4.304} & \textbf{232}\,{\scriptsize\textcolor{green!50!black}{(\textbf{1.7$\times$})}} \\
			
			\midrule
			\multicolumn{9}{c}{\textit{Half-Precision Floating-Point (FP16)}} \\
			\midrule
			ViT-L                & 303.4 & 78.4 & 119.0 & 0.742 & 59.0\% & 123.1 & 1.258 & 795 \\
			Naive EQ-ViT-L       & \textbf{75.8}  & \textbf{80.6} & 119.0 & 0.752 & 59.6\% & 123.1 & 1.262 & 792 \\
			\flashrow
			\textbf{Flash EQ-ViT-L}
			& \textbf{75.8} & \textbf{80.6} & \textbf{44.9} & \textbf{0.497} & \textbf{49.5\%}
			& \textbf{49.0} & \textbf{1.004} & \textbf{996}\,{\scriptsize\textcolor{green!50!black}{(\textbf{1.3$\times$})}} \\
			
			ViT-H                & 631.0 & 81.3 & 248.0 & 1.541 & 64.8\% & 254.7 & 2.379 & 420 \\
			Naive EQ-ViT-H       & \textbf{157.7} & \textbf{81.5} & 248.0 & 1.557 & 65.1\% & 254.7 & 2.392 & 418 \\
			\flashrow
			\textbf{Flash EQ-ViT-H}
			& \textbf{157.7} & \textbf{81.5} & \textbf{93.4} & \textbf{0.976} & \textbf{54.2\%}
			& \textbf{100.1} & \textbf{1.801} & \textbf{555}\,{\scriptsize\textcolor{green!50!black}{(\textbf{1.3$\times$})}} \\
			\bottomrule
		\end{tabular}
	}
\end{table*}

\textbf{Experimental results.}
Tables~\ref{tab:flash-eqlinear-forward} and~\ref{tab:flash_eqlinear_backward_fp32} report forward and backward latency across channel widths, respectively.
\textbf{(i) Forward latency.}
As the channel width $\nicefrac{C}{T}$ increases from $64$ to $2048$, Flash EQ-Linear accelerates the forward pass by $\bm{1.6\times}$--$\bm{2.1\times}$ in FP32 and $\bm{1.3\times}$--$\bm{2.1\times}$ in FP16.
At small channel widths, performance is dominated by memory traffic, layout transformations, and kernel-launch overhead. As the workload becomes compute-bound, the speedup increases toward the theoretical upper bound of $\bm{2.67\times}$.
Flash EQ-Linear also reduces the computational cost from $0.134$ G to $0.052$ G FLOPs, a $\bm{2.6\times}$ reduction consistent with our theoretical complexity analysis.
\textbf{(ii) Backward latency.}
In FP32, Flash EQ-Linear consistently accelerates the backward pass over Standard Linear, although by a smaller margin than that in the forward pass because gradient computation incurs additional arithmetic and memory traffic. 
Comparing the two realizations of Flash EQ-Linear, Flash EQ-Linear (CUDA) consistently outperforms its PyTorch counterpart, especially at smaller channel widths and during backpropagation, highlighting the importance of CUDA optimization.
\textbf{Overall.}
These results confirm that the theoretical reduction in computation translates into substantial wall-clock acceleration.
In the compute-bound regime, Flash EQ-Linear achieves up to a $\bm{2.1\times}$ forward speedup over a highly optimized non-equivariant baseline.

\subsection{Network-Level Acceleration}
\label{subsec:speed_vit_swin}

\textbf{Experimental settings.}
We first integrate Flash EQ-Linear into EQ-ViT~\citep{fu2026vanilla} by replacing every naive EQ-Linear layer while leaving the remaining architecture unchanged.
We compare Flash EQ-ViT against the standard non-equivariant ViT~\citep{dosovitskiy2021vit} and its naive equivariant counterpart across multiple model scales (Tiny/Small/Base/Large/Huge).
Within each scale, all methods share the same total channel width, batch size $128$, input resolution $224\times224$, and group size $T=4$.
We report Top-1 accuracy on ImageNet-100, a 100-class subset of ImageNet-1K~\citep{deng2009imagenet}, together with the parameter count, FLOPs, cumulative linear-layer latency, the linear-layer-to-network latency ratio, end-to-end latency, and inference throughput (imgs/s).
To evaluate the plug-and-play applicability, we additionally integrate Flash EQ-Linear into pretrained EQ-Swin-H~\citep{fu2026vanilla}, EQ-VMamba-H~\citep{zhao2026rotation}, and EQ-INR-LIIF/LTE~\citep{xie2025rotation} for $2\times$ super-resolution, without retraining or fine-tuning.
We further compare Flash EQ-ViT with the $\mathcal{D}_8$, $\mathcal{I}_8$, and $\mathcal{H}_8$ variants of Quick ViTs~\citep{nordstrom2026octic} on ImageNet-100 in FP32 at comparable parameter counts.
All measurements are obtained on a single NVIDIA RTX~4090 GPU and averaged over $200$ runs.

\begin{table*}[t]
	\centering
	\caption{
		\textbf{Plug-and-play speedups} achieved by Flash EQ-Linear over naive equivariant implementations of Swin Transformer, VMamba, SwinIR-LIIF, and SwinIR-LTE in FP32.
	}
	\vspace{-2mm}
	\label{tab:flash_EQSwin_inference}
	\setlength{\tabcolsep}{3pt}
	\renewcommand{\arraystretch}{1.12}
	\resizebox{\textwidth}{!}{
		\begin{tabular}{l r c rr cc cc c}
			\toprule
			\multirow{2.5}{*}{{Method}}
			& \multirow{2.5}{*}{\makecell{{\#Param.}\\{(M)}}}
			& \multirow{2.5}{*}{\makecell{{Top-1} \textbf{(\%)} / \\ PSNR (dB)}}
			& \multicolumn{2}{c}{{\makecell{{FLOPs} {(G)}$\downarrow$}}}
			& \multicolumn{2}{c}{{\makecell{{Latency} {(ms)}$\downarrow$}}}
			& \multicolumn{3}{c}{{\makecell{{Throughput} {(imgs/s)}$\uparrow$}}}  \\
			\cmidrule(lr){4-5}
			\cmidrule(lr){6-7}
			\cmidrule(lr){8-10}
			& & & Naive & Flash & Naive & Flash & Naive & Flash & Speedup  \\
			\midrule
			EQ-Swin-H~\citep{fu2026vanilla}        & {164.0} & {88.3} & 230.3 & \textbf{90.8} & 7.396 & \textbf{4.694} & 135 & \textbf{213} & {\textcolor{green!50!black} {\textbf{1.6$\times$}}} \\
			EQ-VMamba-H~\citep{zhao2026rotation}   & {161.9} & {89.2} & 227.3 & \textbf{101.4} & 7.735 & \textbf{5.441} & 129 & \textbf{184} & {\textcolor{green!50!black}{\textbf{1.4$\times$}}} \\
			EQ-INR-LIIF~\citep{xie2025rotation} & 7.8 & 33.54 & 163.6 & \textbf{93.8} & 9.542 & \textbf{7.572} & 105 & \textbf{132} & {\textcolor{green!50!black}{\textbf{1.3$\times$}}} \\
			EQ-INR-LTE~\citep{xie2025rotation}  & 7.7 & 33.67 & 157.2 & \textbf{89.2} & 10.446 & \textbf{8.243} & 96 & \textbf{121} & {\textcolor{green!50!black}{\textbf{1.3$\times$}}} \\
			\bottomrule	
		\end{tabular}
	}
	\vspace{-3mm}
\end{table*}

\begin{table*}[t]
	\centering
	\caption{
		Comparison of Flash EQ-ViT with Quick ViTs on ImageNet-100 under FP32.
	}
	\vspace{-3mm}
	\label{tab:flash_Octic}
	\setlength{\tabcolsep}{2pt}
	\newcommand{\flashrow}{\rowcolor{gray!10}}
	\renewcommand{\arraystretch}{1.12}
	\resizebox{\textwidth}{!}{
		\begin{tabular}{l c c c c c}
			\toprule
			{{Method}}
			& {\makecell{{\#Param.} {(M)}$\downarrow$}}
			& {\makecell{{FLOPs} {(G)}$\downarrow$}}
			& {\makecell{{Top-1} \textbf{(\%)}$\uparrow$}}
			& {\makecell{{Latency} {(ms)}$\downarrow$}}
			& {\makecell{{Throughput} {(imgs/s)}$\uparrow$}}
			\\ 
			\midrule
			Naive EQ-ViT-T              & 5.4 & 9.2 & \textbf{77.3} & {0.359} & {2785} \\
			Quick ViT-$\mathcal{D}_8$-T~\citep{nordstrom2026octic} & 2.8 & 1.7 & 71.6 & 0.304 & {3294} \\
			Quick ViT-$\mathcal{H}_8$-T~\citep{nordstrom2026octic} & 4.3 & 2.3 & 76.0 & 0.305 & 3276 \\
			Quick ViT-$\mathcal{I}_8$-T~\citep{nordstrom2026octic} & 4.5 & 2.3 & 76.9 & 0.307 & 3257 \\
			\flashrow Flash EQ-ViT-T              & 5.4 & 4.0 & \textbf{77.3} & \textbf{0.249} & \textbf{4012} \\
			\bottomrule	
		\end{tabular}
	}
	\vspace{-4mm}
\end{table*}

\textbf{Experimental results.}
Tables~\ref{tab:flash_EQViT_inference}--\ref{tab:flash_Octic} report the inference efficiency of Flash EQ-Linear across different equivariant architectures.
\textbf{(i) EQ-ViT speedup.}
Flash EQ-ViT achieves end-to-end speedups of $\bm{1.4\times}$--$\bm{1.7\times}$ in FP32 and approximately $\bm{1.3\times}$ in FP16 over standard ViT baselines, while preserving the Top-1 accuracy of its naive equivariant counterpart at every scale.
The gains grow with model scale, since wider models devote a larger fraction of their runtime to linear layers ($58.4\%$ for ViT-S versus $77.7\%$ for ViT-H) and their larger channel widths push the operator further into the compute-bound regime.
\textbf{(ii) Plug-and-play speedup.}
Beyond ViT, Flash EQ-Linear yields end-to-end speedups of $\bm{1.6\times}$ on EQ-Swin-H, $\bm{1.4\times}$ on EQ-VMamba-H, and $\bm{1.3\times}$ on EQ-INR-LIIF/LTE, respectively, with accuracy and PSNR unchanged.
\textbf{(iii) Comparison with architecture-level acceleration.}
At the Tiny scale, Flash EQ-ViT is $\bm{1.22\times}$ faster than the fastest Quick ViT variant at comparable or higher Top-1 accuracy, which we attribute mainly to implementation (fused kernels versus a PyTorch implementation).
Notably, Quick ViTs obtain efficiency through architectural redesign and require training from scratch, whereas Flash EQ-Linear accelerates pretrained equivariant models with no retraining, no architectural change, and no accuracy trade-off.

\section{Conclusion}

In this paper, we presented Flash EQ-Linear, an exact acceleration algorithm with dedicated CUDA kernels that reduces the cost of an EQ-Linear layer from $NC^{2}$ to $\nicefrac{2(T-1)NC^{2}}{T^{2}}$ MACs.
At the operator level, Flash EQ-Linear achieves up to $\bm{2.1\times}$ forward speedup over PyTorch's \texttt{F.linear}; at the network level, Flash EQ-ViT achieves up to $\bm{1.7\times}$ end-to-end speedup over both equivariant and non-equivariant baselines, with no retraining, no architectural change, and no accuracy trade-off.
Crucially, while equivariant architectures like EQ-ViT and EQ-VMamba already excel in highly symmetric domains such as remote sensing and medical imaging, their deployment has historically been bottlenecked by slow execution.
By resolving this bottleneck, Flash EQ-Linear unlocks the application potential of equivariant networks in real-world, latency-sensitive scenarios, demonstrating that equivariance can simultaneously deliver parameter efficiency and hardware-aware acceleration.

\textbf{Limitations and future work.} 
Our current CUDA implementation focuses on the widely used $\mathrm{p}4$ rotation group. 
A natural extension is to generalize the kernels to other cyclic groups, such as the reflection group.
Moreover, while this work targets the fundamental equivariant linear layer, the same algorithmic principle may extend to other equivariant operators, such as equivariant convolutions, enabling more comprehensive acceleration of equivariant networks.

\bibliography{references}
\bibliographystyle{arxiv_preprint}

\appendix

\clearpage

\section{Appendix}
This appendix provides a detailed complexity analysis of Flash EQ-Linear (Sec.~\ref{app:appendix_more_Complexity_analysis}), its relation to isotypic-basis equivariant architectures (Sec.~\ref{app:relation_isotypic}), CUDA kernel implementation details (Sec.~\ref{sec:appendix_more_cuda_kernel}), and additional experimental results (Sec.~\ref{sec:appendix_more_results}).
The additional experiments cover extended network-level benchmarks, numerical equivalence and equivariance verification, the scaling of the operator-level speedup with the group size ($T \! \in \! \{2, 4, 8\}$), and comparisons with Quick ViTs.

\subsection{Detailed Complexity Analysis of Flash EQ-Linear}
\label{app:appendix_more_Complexity_analysis}

This section derives the per-step MAC counts in Theorem~\ref{theorem:MACs_Flash_EQ-Linear}.
Following Sec.~\ref{subsec:Algorithm_Complexity}, we consider an EQ-Linear layer applied to $N$ tokens, with equal input and output channel widths, an even group size $T$, and $T$ dividing $C$.
We count real-valued multiply--accumulate operations, excluding standalone additions, subtractions, and fixed power-of-two rescalings.

\textbf{MACs of Step~1: Group-wise DFT.}
The group-wise DFT is a complex matrix multiplication along the group dimension, $\hat{\bm X}=  {\bm X} \cdot {\bm F} $, where ${\bm F}\in\mathbb{C}^{T\times T}$ is the DFT matrix.
Computing the $T/2+1$ non-redundant frequency components therefore amounts to multiplying ${\bm X}\in\mathbb{R}^{N\times \frac{C}{T}\times T}$ by the corresponding columns of ${\bm F}$:
\begin{equation}\label{eq:step1_macs}
	\hat{\bm X}_{n, k} = {\bm X}_n \cdot {\bm F}_{k}, \quad \mbox{for}~~ k=0,\ldots,\frac{T}{2}, ~~\mbox{and}~~  n = 0, \ldots, N-1, 
\end{equation}
The DFT coefficients at the zero and Nyquist frequencies ($k=0,T/2$) are real, so these two components involve real-by-real MACs, while the remaining $T/2-1$ non-redundant components involve real-by-complex MACs.
Counting these as one and two real-valued MACs, respectively, gives an upper bound of $2NC+2\frac{NC}{T}T(T/2-1)=NCT$ MACs.
Similarly, transforming $\tilde{\bm W}\in\mathbb{R}^{\frac{C}{T}\times\frac{C}{T}\times T}$ requires $C^2$ MACs.
Step~1 therefore costs at most $NTC+C^2$ MACs.

\textbf{MACs of Step~2: Per-frequency matrix multiplication (dominant).}
At each of the $T/2+1$ non-redundant frequencies, Step~2 applies a channel-wise linear projection to each token:
\begin{equation}\label{eq:step2_macs}
	\hat{\bm Y}_n^{G_k} = \hat{\bm W}^{G_k}\cdot\hat{\bm X}_n^{G_k}, \quad \mbox{for}~~ k=0,\ldots,\frac{T}{2}, ~~\mbox{and}~~  n=0, \ldots, N-1, 
\end{equation}
which maps $\frac{C}{T}$ input channels to $\frac{C}{T}$ output channels, and therefore performs $\frac{NC^2}{T^2}$ multiplications at that frequency.
The self-conjugate frequencies $k=0$ and $k=T/2$ are purely real and each require one real multiplication, whereas the remaining $T/2-1$ are complex and need four real multiplications\footnote{Computing a complex product directly requires 4 real multiplications: $(a  +  b\mathrm{i})(c  + d\mathrm{i}) \! =  \! (ac  - bd) \!+ \! (ad +  bc)\mathrm{i}$.}.
Step~2 therefore costs $\frac{2NC^2}{T^2} + 4(T/2-1)\frac{NC^2}{T^2} =\frac{2(T-1)NC^2}{T^2}$ real-valued MACs, which dominates the overall cost of Flash EQ-Linear.

\textbf{MACs of Step~3: Conjugate-symmetric recovery.}
The remaining frequency components of $\hat{\bm Y}$ are recovered by conjugate copying according to Eq.~\ref{eq:conj_sym}.
This step therefore incurs zero MACs.

\textbf{MACs of Step~4: Group-wise IDFT.}
Similar to Step~1, the group-wise IDFT can be written as ${\bm Y}_n=  \hat{\bm Y}_n  \cdot {\bm F}^{-1}$, where ${\bm F}^{-1}\in\mathbb{C}^{T\times T}$ is the inverse DFT matrix.
A direct evaluation involves $NCT$ complex multiplications, corresponding to at most $4NCT$ real-valued MACs.
Fortunately, the conjugate symmetry $\hat{\bm Y}_n^{G_{T-k}}=\overline{\hat{\bm Y}_n^{G_k}}$ allows each pair of conjugate terms to be combined into a real-valued expression.
For each output group index $t$,  $k=1,\ldots,T/2-1$, and $n = 0, \ldots, N-1$, 
\begin{equation}\label{eq:appendix_conj_pair_idft}
	\small
	\begin{aligned}
		\hat{\bm Y}_n^{G_k} {\bm F}^{-1}_{t,k}  \! +  \hat{\bm Y}_n^{G_{T \! - \! k}}  \!{\bm F}^{-1}_{t,T \! - \! k} 
		\! = \!  \hat{\bm Y}_n^{G_k} {\bm F}^{-1}_{t,k} \!  +  \! \overline{ \hat{\bm Y}_n^{G_k} {\bm F}^{-1}_{t,k}} 
		\! = \!  2\, \mathrm{Re}(\hat{\bm Y}_n^{G_k})   \mathrm{Re}({\bm F}^{-1}_{t,k})  \! -  \!  2\, \mathrm{Im}(\hat{\bm Y}_n^{G_k})  \mathrm{Im}({\bm F}^{-1}_{t,k}),
	\end{aligned}
\end{equation}
Each pair thus requires two real-valued MACs per output scalar instead of eight.
The two self-conjugate frequencies, $k=0$ and $k=T/2$, each contribute one real-valued MAC.
Across all $NC$ output scalars, Step~4 therefore requires at most $2NC+2(T/2-1)NC=NCT$ MACs.

\textbf{Total MACs.}
Combining the four steps, the total computational cost of Flash EQ-Linear is
\begin{equation}\label{eq:total_complexity}
	\mathrm{MACs}_{\text{Flash}} 
	= \underbrace{NCT+C^2}_{\text{group-wise DFT (Step 1)}}
	 +  \underbrace{\frac{2(T-1)NC^2}{T^2}}_{\text{Complex matmul (Step 2)}}
	 +  \underbrace{NCT}_{\text{group-wise IDFT (Step 4)}}.
\end{equation}
In typical equivariant networks, where $N\gg T$ and $C\gg T$, the dominant term is
\begin{equation}
	\mathrm{MACs}_{\text{Flash}}  \approx \frac{2(T-1)NC^2}{T^2},
\end{equation}
while the group-wise DFT and IDFT contribute only lower-order overhead.

For the common settings $T=2$ and $T=4$, the DFT and IDFT simplify to additions, subtractions, and fixed power-of-two rescalings.
Under our counting convention, Steps~1 and~4 therefore incur no additional MACs, leaving Step~2 as the sole contribution.
The resulting counts are $NC^2/2$ and $3NC^2/8$, respectively.

\textbf{Speedup analysis.}
Under equal input and output widths, both naive EQ-Linear and a standard non-equivariant linear layer of the same total width compute a dense $C\times C$ projection and therefore require $NC^2$ MACs. 
When per-frequency matrix multiplication dominates the computational cost, Flash EQ-Linear yields a theoretical speedup of
\begin{equation}
	\label{eq:appendix_speedup}
	\mathrm{Speedup}
	=
	\frac{\mathrm{MACs}_{\mathrm{Naive}}}
	{\mathrm{MACs}_{\mathrm{Flash}}}
	\approx
	\frac{NC^2}{2(T-1)NC^2/T^2} = \frac{T^2}{2(T-1)},
\end{equation}
which gives $16/6\approx\bm{2.67\times}$ for the $\mathrm{p}4$ rotation group ($T=4$). This proves Theorem~\ref{theorem:MACs_Flash_EQ-Linear}.

\subsection{Relation to Isotypic-Basis Equivariant Architectures}
\label{app:relation_isotypic}

\textbf{Shared algebraic foundation.}
The acceleration principle of Sec.~\ref{subsec:Acceleration_Principle} is an instance of a classical fact from the representation theory of finite groups: by Schur's lemma, an equivariant linear map between (multiples of) the regular representation becomes block-diagonal once features are expressed in the isotypic basis, i.e., grouped by irreducible representations~\citep{serre1977linear}.
For cyclic groups, this change of basis is exactly the group-wise DFT.
For $\mathrm{p}4$, the real transform in Eq.~\ref{eq:DFT_p4} decomposes the regular representation of the rotation group into the trivial representation (first row), the sign representation (last row), and the two-dimensional rotation representation (middle rows).
Since the latter is of complex type, its equivariant maps form complex $\frac{C}{T}\times\frac{C}{T}$ matrices, which is precisely the structure of Eq.~\ref{eq:appendix_p4_step2} and accounts for the $\frac{6}{16}NC^2$ MACs of Theorem~\ref{theorem:MACs_Flash_EQ-Linear}.
The same principle underlies \citet{bokman2025flopping}, who decompose features into symmetric and antisymmetric components under horizontal mirroring ($2\times$ fewer FLOPs), and \citet{nordstrom2026octic}, who extend it to the octic group $\mathrm{D}_8$, whose two-dimensional irreducible representation is of real type ($5.33\times$ fewer FLOPs; see also the comparison with Quick ViTs in Appendix~\ref{sec:appendix_more_results}).
In all three cases, the attainable reduction is determined by the irreducible decomposition of the group rather than by the specific architecture.

\textbf{Network-level versus layer-level use.}
The works differ in \emph{where} the change of basis is applied.
\citet{bokman2025flopping} and \citet{nordstrom2026octic} apply it at the network level: features are kept in the isotypic basis throughout the network, pointwise nonlinearities are evaluated by transforming to the spatial basis and back~\citep{bokman2025flopping}, and the resulting architectures are trained from scratch.
Flash EQ-Linear applies it at the layer level: each EQ-Linear layer keeps the regular-representation interface and changes basis internally, so that the weights, inputs, and outputs of existing networks remain unchanged.
The two routes are therefore complementary rather than competing: the network-level route avoids per-layer transforms at the cost of redesign and retraining, whereas the layer-level route accelerates existing models without either, and the fused-kernel techniques developed here could equally benefit the linear layers of isotypic-basis architectures.

\subsection{Detailed CUDA Kernel Implementations}
\label{sec:appendix_more_cuda_kernel}

A direct PyTorch realization of Alg.~\ref{alg:flash_eqlinear} issues the group-wise DFT, the frequency-domain matrix multiplication, and the group-wise IDFT as three separate operators.
Each operator launches its own kernel and writes a full-size tensor to GPU device memory, so the activations are read and written three times per call.
This overhead is most severe for the $\mathrm{p}4$ group: its DFT and IDFT involve only additions, subtractions, and dyadic scalings (Eqs.~\ref{eq:DFT_p4} and~\ref{eq:appendix_IDFT}), so their runtime is dominated by memory access rather than arithmetic and can offset the savings of Step~2.
We therefore fuse Steps~1--4 into a single tiled matrix-multiplication kernel, in which the DFT and IDFT are applied as a change of basis to each entry when it is loaded and stored, rather than executed as standalone operators.

\textbf{GPU terminology.}
We follow the standard CUDA memory hierarchy~\citep{hwu2022programming}.
\emph{Device memory} is the off-chip GPU DRAM that stores all PyTorch tensors; it is visible to every kernel but slow to access.
\emph{Shared memory} is a small on-chip scratchpad allocated per CUDA block and shared by the threads of that block; it is unrelated to host--device unified memory.
\emph{Registers} store the local scalar variables of a CUDA thread and are the fastest storage on the GPU.
\emph{Tiling} partitions a matrix multiplication into fixed-size sub-matrices (\emph{tiles}) that are copied from device memory into shared memory, so that the inner loop accesses only on-chip memory.
A \emph{kernel launch} is one host-side invocation of a CUDA kernel over a grid of threads; its fixed cost is significant when the workload per launch is small.

\textbf{Fused forward kernel.}
We flatten the batch and token dimensions into $N\!=\!BL$ rows and write $C'\!=\!\nicefrac{C}{T}$ for the per-group widths.
Each CUDA thread computes the four group components of one output entry ${\bm Y}_{i,j}\!\in\!\mathbb{R}^{4}$ (row $i$, output channel $j$) by traversing the input channels tile by tile, where each tile $\mathcal{P}$ is a group of input channels that is loaded into shared memory at once.
The kernel consists of three stages that correspond to the steps of Alg.~\ref{alg:flash_eqlinear}.

{(i) Input transform (Step~1).}
When a tile of ${\bm X}$ is loaded, Eq.~\ref{eq:DFT_p4} is applied to each entry, producing the four real coordinates $\hat{x}^{G_0}_{ic}$, $\hat{x}^{G_1}_{\mathrm{Re},ic}$, $\hat{x}^{G_2}_{ic}$, and $\hat{x}^{G_1}_{\mathrm{Im},ic}$, i.e., the entries of $\hat{\bm X}^{G_0}$, $\hat{\bm X}^{G_1}_{\mathrm{Re}}$, $\hat{\bm X}^{G_2}$, and $\hat{\bm X}^{G_1}_{\mathrm{Im}}$.
Since the transform matrix contains only $0$ and $\pm 1$, this stage reduces to four add/subtract expressions; neither an FFT routine nor a complex tensor is used.

{(ii) Frequency-domain accumulation (Step~2).}
With $\hat{w}^{G_k}_{jc}$ denoting the entries of the cached weight planes, four accumulators held in registers collect the products of Eq.~\ref{eq:appendix_p4_step2} over all tiles:
\begin{equation}\label{eq:appendix_kernel_mainloop}
	\small
	\begin{aligned}
		a^{G_0}_{ij} &= \sum_{\mathcal{P}} \sum_{c\in\mathcal{P}} \hat{x}^{G_0}_{ic}\,\hat{w}^{G_0}_{jc},
		\qquad\qquad
		a^{G_2}_{ij} = \sum_{\mathcal{P}} \sum_{c\in\mathcal{P}} \hat{x}^{G_2}_{ic}\,\hat{w}^{G_2}_{jc}, \\
		a^{G_1}_{\mathrm{Re},ij} &= \sum_{\mathcal{P}} \sum_{c\in\mathcal{P}} \bigl( \hat{x}^{G_1}_{\mathrm{Re},ic}\,\hat{w}^{G_1}_{\mathrm{Re},jc} - \hat{x}^{G_1}_{\mathrm{Im},ic}\,\hat{w}^{G_1}_{\mathrm{Im},jc} \bigr), \\
		a^{G_1}_{\mathrm{Im},ij} &= \sum_{\mathcal{P}} \sum_{c\in\mathcal{P}} \bigl( \hat{x}^{G_1}_{\mathrm{Re},ic}\,\hat{w}^{G_1}_{\mathrm{Im},jc} + \hat{x}^{G_1}_{\mathrm{Im},ic}\,\hat{w}^{G_1}_{\mathrm{Re},jc} \bigr).
	\end{aligned}
\end{equation}
Eq.~\ref{eq:appendix_kernel_mainloop} is Eq.~\ref{eq:appendix_p4_step2} written entry-wise.
All six real products are accumulated in a single pass over the input channels and share one load of each tile, instead of being executed as six separate GEMM calls on intermediate tensors.

{(iii) Output transform (Steps~3--4).}
The redundant component $\hat{\bm Y}^{G_3}$ is never formed.
The inverse transform of Eq.~\ref{eq:appendix_IDFT}, which uses only the constants $\{0,\pm\tfrac14,\pm\tfrac12\}$, is applied to the four accumulators, and the result plus the bias $\tilde{b}_j$ is written directly into ${\bm Y}_{i,j}$.

Consequently, neither the unrolled matrix ${\bm W}$ nor the full intermediate tensors $\hat{\bm X}$ and $\hat{\bm Y}$ are materialized in global memory; only ${\bm X}$, the cached weight planes, and ${\bm Y}$ reside there.
The kernel preserves the mathematical computation in Sec.~\ref{subsec:Algorithm_Complexity} while reorganizing its execution and memory access.

\textbf{Memory layout and weight caching.}
Activations keep the external layout $[B, L, C', T]$ with the group dimension contiguous, so the four group components of each entry are adjacent in memory and are read in a single access before being transformed.
The four weight planes are computed once from $\tilde{\bm W}$ by Eq.~\ref{eq:DFT_p4} and cached in device memory, so Step~1 on the weight side is amortized over all rows and calls; the cache is invalidated whenever the parameters are updated in place.
No packed copy of the activations is allocated.
Under FP16, tiles are held in half precision, whereas the accumulators are kept in FP32 to limit round-off error.

\begin{table}[t]
	\centering
	\caption{
		\textbf{Shape-specialized backends of Flash EQ-Linear} for the $\mathrm{p}4$ group.
		All backends compute Eqs.~\ref{eq:DFT_p4}, \ref{eq:appendix_kernel_mainloop}, and~\ref{eq:appendix_IDFT} and agree up to floating-point round-off; they differ in which routine executes the matrix multiplication, how kernels are launched, and how tiles are staged.
	}
	\vspace{-3mm}
	\label{tab:appendix_kernel_variants}
	\small
	\setlength{\tabcolsep}{4pt}
	\renewcommand{\arraystretch}{1.15}
	\begin{tabular}{@{}l >{\raggedright\arraybackslash}p{0.19\linewidth} >{\raggedright\arraybackslash}p{0.40\linewidth} >{\raggedright\arraybackslash}p{0.16\linewidth}@{}}
		\toprule
		Backend & Multiplication routine & Specialization & Target regime \\
		\midrule
		Tiled            & Custom tiled kernel & One kernel launch per call; tiles in shared memory, accumulators in registers & Small $C'$, short sequences \\
		Tiled-persistent & Custom tiled kernel & Kernel instances launched once and reused over a queue of output tiles, removing per-call launch overhead & Small to medium $C'$ \\
		Tiled-static     & Custom tiled kernel & Shapes fixed at compile time to remove boundary checks; group components loaded in vectorized form through two staging buffers & Medium to large $C'$ \\
		CUTLASS          & CUTLASS GEMM templates & Tensor layouts expressed with CuTe; tiles staged by asynchronous copies & Large $C'$ \\
		\bottomrule
	\end{tabular}
\end{table}

\textbf{Shape-specialized backends.}
Different shapes $(B, L, C', C')$ expose different bottlenecks.
At small widths, kernel-launch overhead and partially filled tiles dominate the runtime, whereas at large widths arithmetic throughput dominates, so no single tile size is fastest across the whole range.
We therefore implement four backends (Table~\ref{tab:appendix_kernel_variants}) that compute the same quantities as the fused kernel above but differ in multiplication routine, launch strategy, and tile staging, together with per-shape adjustments such as tile padding for narrow output widths $C'$.
Each backend is compiled for a grid of tile sizes and thread-grid configurations, so the number of specialized kernels exceeds the backends listed.

\textbf{Forward and backward passes.}
The forward pass uses the full backend family described above.
The backward pass expresses the output gradient $\nabla{\bm Y}$ in the same four frequency coordinates, forms the transposed products of Eq.~\ref{eq:appendix_p4_step2}, and maps the results back through the fixed transforms, yielding $\nabla{\bm X}$, $\nabla\tilde{\bm W}$, and $\nabla\tilde{\bm b}$ in the original parameterization rather than in any backend-specific layout.
Because these gradients do not depend on which forward backend was used, a single fused backward kernel per precision is correct for every forward backend, and we share one such kernel across all of them instead of specializing the backward pass per shape.
This choice is deliberate: forward latency dominates the inference setting that Flash EQ-Linear targets, and a single backward kernel allows the entire gradient path to be verified against the naive implementation in one comparison (Table~\ref{tab:flash_eqlinear_Precision_Errors}).
Even without specialization, the shared backward kernel is faster than both baselines at every channel width (Table~\ref{tab:flash_eqlinear_backward_fp32}); per-shape specialization of the backward pass is left for future work.

\subsection{More Experimental Results}
\label{sec:appendix_more_results}

\begin{table*}[t]
	\centering
	\caption{
		Configuration details, parameter counts, and total FLOPs of the non-equivariant ViT and Swin Transformer baselines used in our experiments.
		Their equivariant counterparts use the same total channel widths at each model scale.
	}
	\vspace{-2mm}
	\label{tab:appendix_vit-swin-configs}
	\begin{tabular}{
			l
			@{\hspace{6.0pt}} l
			@{\hspace{4.0pt}} c
			@{\hspace{3.0pt}} c
			@{\hspace{8.0pt}} c
			@{\hspace{8.0pt}} c
			@{\hspace{8.0pt}} c
			@{\hspace{8.0pt}} c
		}		\toprule
		Model & Scale & \makecell{Patch\\ Size}  & \makecell{Embedding\\Dimension} & \makecell{Depth/Depths}  & \makecell{Num. Heads}  
		& \makecell{\#Param.\\(M)$\downarrow$} 
		& \makecell{FLOPs\\(G)$\downarrow$} \\
		\midrule
		\multirow{5}{*}{ViT}    
		& Tiny   & 16 & 384  & 12 & 3  & 21.7  & 9.2 \\
		& Small  & 16 & 480  & 12 & 6  & 33.8  & 14.1 \\
		& Base   & 16 & 768  & 12 & 12 & 85.9  & 35.1 \\
		& Large  & 16 & 1024 & 24 & 16 & 303.4 & 123.1 \\
		& Huge   & 16 & 1280 & 32 & 16 & 631.0 & 254.7 \\
		\midrule
		\multirow{5}{*}{Swin}   
		& Tiny   & 4 & $[96,192,384,768]$  & $[2,2,6,2]$  & $[3,6,12,24]$   & 27.6  & 8.99 \\
		& Small  & 4 & $[96,192,384,768]$  & $[2,2,18,2]$ & $[3,6,12,24]$   & 48.9  & 17.5 \\
		& Base   & 4 & $[128,256,512,1024]$ & $[2,2,18,2]$ & $[4,8,16,32]$   & 86.8  & 30.9 \\
		& Large  & 4 & $[192,384,768,1536]$ & $[2,2,18,2]$ & $[6,12,24,48]$  & 195.1 & 69.0 \\
		& Huge   & 4 & $[352,704,1408,2816]$ & $[2,2,18,2]$ & $[11,22,44,88]$ & 655.1 & 230.3 \\
		\bottomrule
	\end{tabular}
	\vspace{-2mm}
\end{table*}

\begin{table*}[!t]
	\centering
	\caption{
		\textbf{Inference efficiency comparison} of Flash EQ-ViT against standard non-EQ and naive EQ baselines on ImageNet-100.
		\textit{Speedup ratios} (\textcolor{green!50!black}{green}) are measured against standard baselines.
	}
	\vspace{-1mm}
	\label{tab:appendix_ViT}
	\newcommand{\NA}{--}
	\newcommand{\flashrow}{\rowcolor{gray!10}}
	\setlength{\tabcolsep}{5.5pt}
	\renewcommand{\arraystretch}{1.12}
	\resizebox{\textwidth}{!}{
		\begin{tabular}{l c c l l l l l l}
			\toprule
			\multirow{3}{*}{\textbf{Method}}
			& \multirow{3}{*}{\makecell{\textbf{\#Param.}\\\textbf{(M)}$\downarrow$}}
			& \multirow{3}{*}{\makecell{\textbf{Top-1}\\\textbf{(\%)}$\uparrow$}}
			& \multicolumn{3}{c}{\textbf{All Linear Layers}}
			& \multicolumn{3}{c}{\textbf{Total Network}} \\
			\cmidrule(lr){4-6}
			\cmidrule(lr){7-9}
			&
			&
			& \makecell{\textbf{FLOPs}\\\textbf{(G)}$\downarrow$}
			& \makecell{\textbf{Latency}\\\textbf{(ms)}$\downarrow$}
			& \makecell{\textbf{Latency}\\\textbf{Ratio}}
			& \makecell{\textbf{FLOPs}\\\textbf{(G)}$\downarrow$}
			& \makecell{\textbf{Latency}\\\textbf{(ms)}$\downarrow$}
			& \makecell{\textbf{Throughput}\\\textbf{(imgs/s)}$\uparrow$} \\
			\midrule
			\multicolumn{9}{c}{\textit{Single-Precision Floating-Point (FP32)}} \\
			\midrule
			ViT-T                & 21.7  & 76.5 & 8.4 & 0.218 & 60.8\% & 9.2 & 0.358  & 2796 \\
			Naive EQ-ViT-T       & \textbf{5.4}  & \textbf{77.3} & 8.4 & 0.217 & 60.3\% & 9.2 & 0.359  & 2785 \\
			\flashrow
			\textbf{Flash EQ-ViT-T} & \textbf{5.4} & \textbf{77.3} & \textbf{3.2} & \textbf{0.115} & \textbf{46.2\%}
			& \textbf{4.0} & \textbf{0.249} & \textbf{4012}\,{\scriptsize\textcolor{green!50!black}{(\textbf{1.4$\times$})}} \\
			ViT-S                 & 33.8 & 76.5 & 13.1 & 0.352 & 58.4\% & 14.1 & 0.603  & 1658 \\
			Naive EQ-ViT-S        & \textbf{8.4}  & \textbf{78.2} & 13.1 & 0.355 & 58.7\% & 14.1 & 0.604 & 1656 \\
			\flashrow
			\textbf{Flash EQ-ViT-S}
			& \textbf{8.4} & \textbf{78.2} & \textbf{5.0} & \textbf{0.183} & \textbf{42.6\%}
			& \textbf{6.0} & \textbf{0.429} & \textbf{2332}\,{\scriptsize\textcolor{green!50!black}{(\textbf{1.4$\times$})}} \\
			ViT-B                 & 85.9 & 77.1 & 33.5 & 0.761 & 67.1\% & 35.1 & 1.135 & 881 \\
			Naive EQ-ViT-B        & \textbf{21.4} & \textbf{80.1} & 33.5 & 0.757 & 67.1\% & 35.1 & 1.128 & 886 \\
			\flashrow
			\textbf{Flash EQ-ViT-B}
			& \textbf{21.4} & \textbf{80.1} & \textbf{12.6} & \textbf{0.400} & \textbf{52.3\%}
			& \textbf{14.3} & \textbf{0.764} & \textbf{1309}\,{\scriptsize\textcolor{green!50!black}{(\textbf{1.5$\times$})}} \\
			ViT-L                 & 303.4 & 78.4 & 119.0 & 2.672 & 71.9\% & 123.1 & 3.715 & 269 \\
			Naive EQ-ViT-L        & \textbf{75.8} & \textbf{80.6} & 119.0 & 2.760 & 73.9\% & 123.1 & 3.735 & 268 \\
			\flashrow
			\textbf{Flash EQ-ViT-L}
			& \textbf{75.8} & \textbf{80.6} & \textbf{44.9} & \textbf{1.336} & \textbf{56.7\%}
			& \textbf{49.0} & \textbf{2.356} & \textbf{424}\,{\scriptsize\textcolor{green!50!black}{(\textbf{1.6$\times$})}} \\
			ViT-H                 & 631.0 & 81.3 & 248.0 & 5.586 & 77.7\% & 254.7 & 7.192 & 139 \\
			Naive EQ-ViT-H        & \textbf{157.7} & \textbf{81.5} & 248.0 & 5.627 & 78.4\% & 254.7 & 7.182 & 139 \\
			\flashrow
			\textbf{Flash EQ-ViT-H}
			& \textbf{157.7} & \textbf{81.5} & \textbf{93.4} & \textbf{2.719} & \textbf{63.2\%}
			& \textbf{100.1} & \textbf{4.304} & \textbf{232}\,{\scriptsize\textcolor{green!50!black}{(\textbf{1.7$\times$})}} \\
			
			\midrule
			\multicolumn{9}{c}{\textit{Half-Precision Floating-Point (FP16)}} \\
			\midrule
			ViT-T                & 21.7  & 76.5 & 8.4 & 0.074 & 57.4\% & 9.2 & 0.128  & 7794 \\
			Naive EQ-ViT-T       & \textbf{5.4}  & \textbf{77.3} & 8.4 & 0.075 & 57.3\% & 9.2 & 0.130  & 7685 \\
			\flashrow
			\textbf{Flash EQ-ViT-T} & \textbf{5.4} & \textbf{77.3} & \textbf{3.2} & \textbf{0.058} & \textbf{51.3\%}
			& \textbf{4.0} & \textbf{0.113} & \textbf{8848}\,{\scriptsize\textcolor{green!50!black}{(\textbf{1.1$\times$})}} \\
			
			ViT-S                & 33.8 & 76.5 & 13.1 & 0.105 & 52.6\% & 14.1 & 0.199 & 5030 \\
			Naive EQ-ViT-S       & \textbf{8.4} & \textbf{78.2} & 13.1 & 0.109 & 54.5\% & 14.1 & 0.200 & 5010 \\
			\flashrow
			\textbf{Flash EQ-ViT-S}
			& \textbf{8.4} & \textbf{78.2} & \textbf{5.0} & \textbf{0.084} & \textbf{48.4\%}
			& \textbf{6.0} & \textbf{0.174} & \textbf{5750}\,{\scriptsize\textcolor{green!50!black}{(\textbf{1.1$\times$})}} \\
			
			ViT-B                & 85.9 & 77.1 & 33.5 & 0.215 & 57.6\% & 35.1 & 0.372 & 2688 \\
			Naive EQ-ViT-B       & \textbf{21.4} & \textbf{80.1} & 33.5 & 0.218 & 58.3\% & 35.1 & 0.374 & 2676 \\
			\flashrow
			\textbf{Flash EQ-ViT-B}
			& \textbf{21.4} & \textbf{80.1} & \textbf{12.6} & \textbf{0.147} & \textbf{49.1\%}
			& \textbf{14.3} & \textbf{0.300} & \textbf{3330}\,{\scriptsize\textcolor{green!50!black}{(\textbf{1.2$\times$})}} \\
			
			ViT-L                & 303.4 & 78.4 & 119.0 & 0.742 & 59.0\% & 123.1 & 1.258 & 795 \\
			Naive EQ-ViT-L       & \textbf{75.8}  & \textbf{80.6} & 119.0 & 0.752 & 59.6\% & 123.1 & 1.262 & 792 \\
			\flashrow
			\textbf{Flash EQ-ViT-L}
			& \textbf{75.8} & \textbf{80.6} & \textbf{44.9} & \textbf{0.497} & \textbf{49.5\%}
			& \textbf{49.0} & \textbf{1.004} & \textbf{996}\,{\scriptsize\textcolor{green!50!black}{(\textbf{1.3$\times$})}} \\
			
			ViT-H                & 631.0 & 81.3 & 248.0 & 1.541 & 64.8\% & 254.7 & 2.379 & 420 \\
			Naive EQ-ViT-H       & \textbf{157.7} & \textbf{81.5} & 248.0 & 1.557 & 65.1\% & 254.7 & 2.392 & 418 \\
			\flashrow
			\textbf{Flash EQ-ViT-H}
			& \textbf{157.7} & \textbf{81.5} & \textbf{93.4} & \textbf{0.976} & \textbf{54.2\%}
			& \textbf{100.1} & \textbf{1.801} & \textbf{555}\,{\scriptsize\textcolor{green!50!black}{(\textbf{1.3$\times$})}} \\
			\bottomrule
		\end{tabular}
	}
	\vspace{-5mm}
\end{table*}

\textbf{Runtime measurement protocol.}
All timings use CUDA events with synchronization before and after each timed loop, exclude input generation, host--device transfers, and model initialization.
\textbf{(i) Operator benchmarks} are averaged over $200$ iterations after $5$ warm-up iterations. Naive EQ-Linear re-expands its weights on every forward call, and backward timing excludes the forward pass and optimizer updates.
\textbf{(ii) Network benchmarks} run in evaluation mode without gradient tracking and are averaged over $200$ forward passes after $20$ warm-up iterations. Linear-layer latency is profiled separately with CUDA-event hooks, which are removed for end-to-end timing, so the reported latency ratio is a profiling-based estimate. We report per-image latency (batch latency divided by $B$) and the corresponding throughput.
FP16 runs cast parameters and inputs to half precision without automatic mixed precision.

\textbf{Configuration details of ViT and Swin.}
Table~\ref{tab:appendix_vit-swin-configs} lists the architectural configurations of the ViT and Swin Transformer variants~\citep{dosovitskiy2021vit, liu2021swin} used in Sec.~\ref{subsec:speed_vit_swin}.
Each family spans five scales of increasing capacity (Tiny / Small / Base / Large / Huge).
ViT scales the embedding dimension $C$ and the number of attention heads, from $21.7$M parameters at ViT-T to $631.0$M at ViT-H; Swin retains its four-stage hierarchy and scales the per-stage embedding dimensions, depths, and head counts, from $27.6$M at Swin-T to $655.1$M at Swin-H.
The Naive EQ and Flash EQ variants share these configurations exactly but operate on the $\mathrm{p}4$ rotation group ($T \!=\! 4$): the total channel width $C$ is matched to the non-equivariant baseline at every scale, so the per-group width is $\nicefrac{C}{T}$.
This matching is what makes the comparison like-for-like---at a given scale the three variants differ only in how the same $C \!\times\! C$ projection is computed, not in its size.

\begin{table*}[!t]
	\centering
	\caption{
		\textbf{Inference efficiency comparison} of Flash EQ-Swin against standard non-EQ and naive EQ baselines on ImageNet-100.
		\textit{Speedup ratios} (\textcolor{green!50!black}{green}) are measured against standard baselines.
	}
	\vspace{-3mm}
	\label{tab:appendix_Swin}
	\newcommand{\flashrow}{\rowcolor{gray!10}}
	\setlength{\tabcolsep}{5.5pt}
	\renewcommand{\arraystretch}{1.12}
	\resizebox{\textwidth}{!}{
		
		\begin{tabular}{l c c l l l l l l}
			\toprule
			\multirow{3}{*}{\textbf{Method}}
			& \multirow{3}{*}{\makecell{\textbf{\#Param.}\\\textbf{(M)}$\downarrow$}}
			& \multirow{3}{*}{\makecell{\textbf{Top-1}\\\textbf{(\%)}$\uparrow$}}
			& \multicolumn{3}{c}{\textbf{All Linear Layers}}
			& \multicolumn{3}{c}{\textbf{Total Network}} \\
			\cmidrule(lr){4-6}
			\cmidrule(lr){7-9}
			&
			&
			& \makecell{\textbf{FLOPs}\\\textbf{(G)}$\downarrow$}
			& \makecell{\textbf{Latency}\\\textbf{(ms)}$\downarrow$}
			& \makecell{\textbf{Latency}\\\textbf{Ratio}}
			& \makecell{\textbf{FLOPs}\\\textbf{(G)}$\downarrow$}
			& \makecell{\textbf{Latency}\\\textbf{(ms)}$\downarrow$}
			& \makecell{\textbf{Throughput}\\\textbf{(imgs/s)}$\uparrow$} \\
			\midrule
			\multicolumn{9}{c}{\textit{Single-Precision Floating-Point (FP32)}} \\
			\midrule
			
			Swin-T                & 27.6 & 87.1 & 8.68 & 0.291 & 46.1\% & 8.99 & 0.631 & 1585 \\
			Naive EQ-Swin-T       & \textbf{6.9} & \textbf{87.3} & 8.68 & 0.300 & 47.4\% & 8.99 & 0.633 & 1580 \\
			\flashrow
			\textbf{Flash EQ-Swin-T}
			& \textbf{6.9} & \textbf{87.3} & \textbf{3.55} & \textbf{0.142} & \textbf{29.0\%}
			& \textbf{3.85} & \textbf{0.490} & \textbf{2041}\,{\scriptsize\textcolor{green!50!black}{(\textbf{1.3$\times$})}} \\
			
			Swin-S                & 48.9 & 87.6 & 17.0 & 0.513 & 49.3\% & 17.5 & 1.040 & 962 \\
			Naive EQ-Swin-S       & \textbf{12.3} & \textbf{87.7} & 17.0 & 0.527 & 50.3\% & 17.5 & 1.047 & 956 \\
			\flashrow
			\textbf{Flash EQ-Swin-S}
			& \textbf{12.3} & \textbf{87.7} & \textbf{6.72} & \textbf{0.276} & \textbf{34.9\%}
			& \textbf{7.20} & \textbf{0.791} & \textbf{1265}\,{\scriptsize\textcolor{green!50!black}{(\textbf{1.3$\times$})}} \\
			
			Swin-B                & 86.8 & 87.8 & 30.2 & 0.812 & 53.2\% & 30.9 & 1.527 & 655 \\
			Naive EQ-Swin-B       & \textbf{21.8} & \textbf{88.6} & 30.2 & 0.806 & 52.6\% & 30.9 & 1.534 & 652 \\
			\flashrow
			\textbf{Flash EQ-Swin-B}
			& \textbf{21.8} & \textbf{88.6} & \textbf{11.9} & \textbf{0.442} & \textbf{38.4\%}
			& \textbf{12.5} & \textbf{1.151} & \textbf{868}\,{\scriptsize\textcolor{green!50!black}{(\textbf{1.3$\times$})}} \\
			
			Swin-L                & 195.1 & 87.5 & 68.0 & 1.661 & 59.4\% & 69.0 & 2.799 & 357 \\
			Naive EQ-Swin-L       & \textbf{48.9} & \textbf{88.2} & 68.0 & 1.672 & 59.7\% & 69.0 & 2.800 & 357 \\
			\flashrow
			\textbf{Flash EQ-Swin-L}
			& \textbf{48.9} & \textbf{88.2} & \textbf{26.6} & \textbf{0.875} & \textbf{44.1\%}
			& \textbf{27.6} & \textbf{1.983} & \textbf{504}\,{\scriptsize\textcolor{green!50!black}{(\textbf{1.4$\times$})}} \\
			
			Swin-H                & 655.1 & 87.2 & 228.6 & 5.239 & 71.0\% & 230.3 & 7.381 & 135 \\
			Naive EQ-Swin-H       & \textbf{164.0} & \textbf{88.3} & 228.6 & 5.273 & 71.3\% & 230.3 & 7.396 & 135 \\
			\flashrow
			\textbf{Flash EQ-Swin-H}
			& \textbf{164.0} & \textbf{88.3} & \textbf{89.1} & \textbf{2.639} & \textbf{56.2\%}
			& \textbf{90.8} & \textbf{4.694} & \textbf{213}\,{\scriptsize\textcolor{green!50!black}{(\textbf{1.6$\times$})}} \\
			
			\midrule
			\multicolumn{9}{c}{\textit{Half-Precision Floating-Point (FP16)}} \\
			\midrule
			
			Swin-T                & 27.6 & 87.1 & 8.68 & {0.088} & {37.8\%} & 8.99 & 0.232 & 4315 \\
			Naive EQ-Swin-T       & \textbf{6.9} & \textbf{87.3} & 8.68 & 0.091 & 39.1\% & 8.99 & 0.233 & 4292 \\
			\flashrow
			\textbf{Flash EQ-Swin-T}
			& \textbf{6.9} & \textbf{87.3} & \textbf{3.55} & \textbf{0.085} & \textbf{38.5\%}
			& \textbf{3.85} & \textbf{0.220} & \textbf{4554}\,{\scriptsize\textcolor{green!50!black}{(\textbf{1.1$\times$})}} \\
			
			Swin-S                & 48.9 & 87.6 & 17.0 & {0.168} & {46.0\%} & 17.5 & 0.366 & 2733 \\
			Naive EQ-Swin-S       & \textbf{12.3} & \textbf{87.7} & 17.0 & 0.177 & 48.2\% & 17.5 & 0.367 & 2727 \\
			\flashrow
			\textbf{Flash EQ-Swin-S}
			& \textbf{12.3} & \textbf{87.7} & \textbf{6.72} & \textbf{0.150} & \textbf{45.0\%}
			& \textbf{7.20} & \textbf{0.332} & \textbf{3009}\,{\scriptsize\textcolor{green!50!black}{(\textbf{1.1$\times$})}} \\
			
			Swin-B                & 86.8 & 87.8 & 30.2 & {0.238} & {44.9\%} & 30.9 & {0.529} & {1889} \\
			Naive EQ-Swin-B       & \textbf{21.8} & \textbf{88.6} & 30.2 & 0.245 & 45.7\% & 30.9 & 0.535 & 1869 \\
			\flashrow
			\textbf{Flash EQ-Swin-B}
			& \textbf{21.8} & \textbf{88.6} & \textbf{11.9} & \textbf{0.185} & \textbf{39.4\%}
			& \textbf{12.5} & \textbf{0.470} & \textbf{2129}\,{\scriptsize\textcolor{green!50!black}{(\textbf{1.1$\times$})}} \\
			
			Swin-L                & 195.1 & 87.5 & 68.0 & {0.463} & {48.3\%} & 69.0 & {0.958} & {1044} \\
			Naive EQ-Swin-L       & \textbf{48.9} & \textbf{88.2} & 68.0 & 0.469 & 48.8\% & 69.0 & 0.961 & 1041 \\
			\flashrow
			\textbf{Flash EQ-Swin-L}
			& \textbf{48.9} & \textbf{88.2} & \textbf{26.6} & \textbf{0.317} & \textbf{38.8\%}
			& \textbf{27.6} & \textbf{0.817} & \textbf{1224}\,{\scriptsize\textcolor{green!50!black}{(\textbf{1.2$\times$})}} \\
			
			Swin-H                & 655.1 & 87.2 & 228.6 & {1.432} & {59.3\%} & 230.3 & {2.414} & {414} \\
			Naive EQ-Swin-H       & \textbf{164.0} & \textbf{88.3} & 228.6 & 1.445 & 59.8\% & 230.3 & 2.418 & {414} \\
			\flashrow
			\textbf{Flash EQ-Swin-H}
			& \textbf{164.0} & \textbf{88.3} & \textbf{89.1} & \textbf{0.958} & \textbf{49.0\%}
			& \textbf{90.8} & \textbf{1.955} & \textbf{512}\,{\scriptsize\textcolor{green!50!black}{(\textbf{1.2$\times$})}} \\
			\bottomrule
		\end{tabular}
	}
	\vspace{-2mm}
\end{table*}

\textbf{Flash EQ-ViT and Flash EQ-Swin.}
Tables~\ref{tab:appendix_ViT} and~\ref{tab:appendix_Swin} report the complete results for ViT and for Swin respectively, each covering all five scales in both FP32 and FP16.
\textbf{(i)} Flash EQ-Linear reduces the cumulative linear-layer latency by $\sim\! \bm{2 \times}$ over the Naive EQ baseline at every scale in FP32, and by $1.1\times$--$1.6\times$ in FP16, yielding $\bm{1.1\times}$--$\bm{1.7\times}$ end-to-end speedup across the twenty configurations.
The FP16 margin is smaller because half precision roughly doubles arithmetic throughput, leaving the same layer closer to the memory-bound regime; the effect is strongest for Swin, whose windowed attention uses narrower linear layers than ViT at matched total width.
\textbf{(ii)} The end-to-end speedup tracks the fraction of runtime the linear layers occupy rather than model size as such, as established in Sec.~\ref{subsec:speed_vit_swin}.
That fraction generally grows with scale and the measured speedups follow it in both directions, with ViT-T and ViT-S both at $\bm{1.4\times}$ while ViT-H reaches $\bm{1.7\times}$ as its linear layers come to dominate the runtime.
Across all scales and both families, Flash EQ-ViT/Swin reproduces the ImageNet-100 Top-1 accuracy of its Naive EQ counterpart exactly, as guaranteed by the exactness established in Sec.~\ref{subsec:Algorithm_Complexity}.

\begin{table*}[t]
	\centering
	\caption{
		\textbf{Validation of numerical equivalence} between Flash EQ-Linear and Naive EQ-Linear in FP32. 
		The relative $L_2$ error (Rel $L_2$) is the primary metric.
	}
	\vspace{-3mm}
	\label{tab:flash_eqlinear_Precision_Errors}
	\setlength{\tabcolsep}{3.0pt}
	\resizebox{\textwidth}{!}{
		\begin{tabular}{l c c c c c c }
			\toprule
			Tensor & Dtype & Shape & Rel $L_2$ $\downarrow$ & Mean Abs $\downarrow$ & p99 Rel $\downarrow$ & Match\\
			\midrule
			Forward output $\bm Y$                      & FP32 & $\mathrm{ B \!\! \times \!\! L \!\! \times \!\! C/T \!\! \times \!\! T}$  & $2.6 \! \times \! 10^{-7}$ & $1.1 \! \times \! 10^{-7}$ & $1.7 \! \times \! 10^{-6}$  & \checkmark \\
			Backward input grad $\bm \nabla \! \bm X$   & FP32 & $\mathrm{ B \!\! \times \!\! L \!\! \times \!\! C/T \!\! \times \!\! T}$  & $2.2 \! \times \! 10^{-7}$ & $9.4 \! \times \! 10^{-8}$ & $1.1 \! \times \! 10^{-5}$ & \checkmark \\
			Backward weight grad $\bm \nabla \tilde{\bm W}$     & FP32 & $\mathrm{ C/T \!\! \times \!\! C/T \!\! \times \!\! T}$  & $5.9 \! \times \! 10^{-7}$ & $2.9 \! \times \! 10^{-5}$ & $3.2 \! \times \! 10^{-5}$ & \checkmark \\
			Backward bias grad $\bm \nabla \tilde{\bm b}$       & FP32 & $\mathrm{C/T}$   & $1.5 \! \times \! 10^{-7}$ & $7.2 \! \times \! 10^{-6}$ & $4.5 \! \times \! 10^{-6}$ & \checkmark \\
			\bottomrule
		\end{tabular}
	}
\end{table*}

\begin{table*}[t]
	\centering
	\caption{\textbf{Equivariance verification} of Flash EQ-Linear on the $\mathrm{p}4$ rotation group in FP32.}
	\vspace{-3mm}
	\label{tab:eq_error}
	\setlength{\tabcolsep}{6pt}
	\begin{tabular}{l c c c c c}
		\toprule
		Method & NMAE $\downarrow$ &  Max Abs $\downarrow$ & Mean Abs $\downarrow$ & p99 Rel $\downarrow$ & Equivariant \\
		\midrule
		Standard Linear & $1.4 \!\times\! 10^{0}$ & $4.1 \!\times\! 10^{0}$ & $6.5 \!\times\! 10^{-1}$ & $6.4 \!\times\! 10^{1}$  & \ding{55} \\
		Naive EQ-Linear & $1.3 \!\times\! 10^{-7}$ & $7.3 \!\times\! 10^{-7}$ & $6.0 \!\times\! 10^{-8}$ & $7.5 \!\times\! 10^{-6}$  & \checkmark \\
		\rowcolor{gray!10}
		\textbf{Flash EQ-Linear} & $\bm{4.0 \!\times\! 10^{-8}}$ & $\bm{3.2 \!\times\! 10^{-7}}$ & $\bm{1.9 \!\times\! 10^{-8}}$ & $\bm{2.4 \!\times\! 10^{-6}}$  & \checkmark \\
		\bottomrule
	\end{tabular}
\end{table*}

\textbf{Exactness and equivariance verification.}
We provide empirical support for the two properties established in Sec.~\ref{subsec:Algorithm_Complexity}: numerical equivalence to Naive EQ-Linear, and preservation of rotation equivariance.
\textbf{(i) Numerical equivalence.}
We compare Flash EQ-Linear with Naive EQ-Linear on random inputs of shape $(B, L, \nicefrac{C}{T}, T) \! = \! (1, 1024, 64, 4)$ over a complete training step, i.e., for ${\bm Y}$, $\nabla{\bm X}$, $\nabla\tilde{\bm W}$ and $\nabla\tilde{\bm b}$, using the naive results as the reference.
The primary metric is the relative $L_2$ error, i.e., the $L_2$ norm of the difference divided by that of the reference.
We also report the mean absolute error and the $99$-th percentile of the element-wise relative error, computed over entries whose reference magnitude exceeds $10^{-1}$ for ${\bm Y}$ and $10^{-3}$ for the gradients to exclude near-zero references.
A tensor is marked as matching if its relative $L_2$ error is below $10^{-5}$.
As shown in Table~\ref{tab:flash_eqlinear_Precision_Errors}, the relative $L_2$ error lies between $10^{-7}$ and $10^{-6}$ for every tensor, since the two implementations accumulate the same quantity in a different order.
\textbf{(ii) Equivariance preservation.}
Under the $\mathrm{p}4$ rotation group, rotating the input cyclically permutes its group dimension, so an equivariant layer $f$ must satisfy $f(G({\bm X})) \! = \! G(f({\bm X}))$ for every $G \! \in \! \mathcal{G}$.
Following~\citet{xie2025rotation}, we measure the discrepancy between the two sides with the normalized mean absolute error, $\mathrm{NMAE} = \operatorname{mean}|f(G(\bm X)) - G(f(\bm X))| \, / \operatorname{mean}|G(f(\bm X))|$: values of order $1$ indicate that the two sides are unrelated, whereas values near the FP32 unit round-off ($2^{-24} \! \approx \! 6 \! \times \! 10^{-8}$) indicate equivariance up to rounding.
We additionally report the maximum and mean absolute discrepancy and the $99$-th percentile of the relative discrepancy over entries with $|G(f(\bm X))| > 10^{-1}$, and mark a layer as equivariant if its NMAE is below $10^{-5}$, as in (i).
In Table~\ref{tab:eq_error}, Standard Linear yields an NMAE of $\bm{1.4}$, confirming that an unconstrained layer does not commute with the group action. In contrast, Flash EQ-Linear yields $\bm{4.0 \! \times \! 10^{-8}}$, comparable to Naive EQ-Linear ($\bm{1.3 \! \times \! 10^{-7}}$) and more than seven orders of magnitude below Standard Linear, showing that equivariance is preserved up to floating-point rounding.
\begin{table*}[t]
	\centering
	\caption{
		\textbf{Forward latency scaling with the group size $T$.}
		Flash EQ-Linear (PyTorch) is compared against Standard Linear and Naive EQ-Linear baselines for $T \! \in \! \{2, 4, 8\}$ under different channel numbers ($\nicefrac{C}{T} \! = \! 64 \! \to \! 2048$), measured in FP32 with $B \! = \! 32$ and $L \! = \! 1024$.
		\#Param. and FLOPs are reported at a fixed total width $C \! = \! 2048$.
		\textit{Speedup ratios} (\textcolor{green!50!black}{green}; \textcolor{gray}{gray} if ${<}1$) are measured against the standard Linear (PyTorch's \texttt{F.linear}) baseline.
	}
	\vspace{-3mm}
	\label{tab:flash-eqlinear-T-scaling}
	\newcommand{\speedup}[1]{\textcolor{green!50!black}{\textbf{#1$\times$}}}
	\newcommand{\slowdown}[1]{\textcolor{gray}{#1$\times$}}
	\newcommand{\sci}[2]{$#1 \! \times \! 10^{#2}$}
	
	\setlength{\tabcolsep}{5.5pt}
	\small
	\begin{tabular}{lccc ccccc}
		\toprule
		\multirow{2.5}{*}{Method}
		& \multirow{2.5}{*}{\makecell{\#Param.\\(M)$\downarrow$}}
		& \multirow{2.5}{*}{\makecell{FLOPs\\(G)$\downarrow$}}
		& \multicolumn{6}{c}{Forward Latency (ms)$\downarrow$} \\
		\cmidrule(lr){4-9}
		& & & 64 & 128 & 256 & 512 & 1024 & 2048 \\
		\midrule
		\multicolumn{9}{c}{\textit{Group size $T \! = \! 2$ \, (theoretical speedup $\nicefrac{T^2}{2(T-1)} \! = \! \bm{2.00\times}$)}} \\
		\midrule
		Standard Linear & 4.194 & 8.590  & \textbf{0.04} & \textbf{0.11} & \textbf{0.46} & \textbf{1.68} & 6.73 & 27.86 \\
		Naive EQ-Linear & \textbf{2.097} & 8.590 & 0.08 & 0.12 & 0.51 & 1.69 & 6.83 & 28.18 \\
		\rowcolor{gray!10}
		\textbf{Flash EQ-Linear (PyTorch)} & \textbf{2.097} & \textbf{4.301} &  0.17 & 0.34 & 0.80 & 1.98 & \textbf{5.58} & \textbf{17.25} \\
		\rowcolor{gray!10}
		\textit{Speedup/Reduction $\uparrow$} & \textbf{50.0\%} & \textbf{49.9\%} & \slowdown{0.2} & \slowdown{0.3} & \slowdown{0.6} & \slowdown{0.8} & \speedup{1.2} & \speedup{1.6} \\
		
		\midrule
		\multicolumn{9}{c}{\textit{Group size $T \! = \! 4$ \, (theoretical speedup $\nicefrac{T^2}{2(T-1)} \! = \! \bm{2.67\times}$)}} \\
		\midrule
		Standard Linear & 4.194 & 8.590 & \textbf{0.11} & \textbf{0.43} & \textbf{1.52} & 6.05 & 24.48 & 104.96 \\
		Naive EQ-Linear & \textbf{1.049} & 8.590 & \textbf{0.11} & 0.44 & 1.54 & 6.25 & 25.16 & 105.74 \\
		\rowcolor{gray!10}
		\textbf{Flash EQ-Linear (PyTorch)} & \textbf{1.049} & \textbf{3.233} & 0.52 & 0.99 & 2.24 & \textbf{5.82} & \textbf{16.16} & \textbf{51.89} \\
		\rowcolor{gray!10}
		\textit{Speedup/Reduction $\uparrow$} & \textbf{75.0\%} & \textbf{62.4\%} & \slowdown{0.2} & \slowdown{0.4} & \slowdown{0.7} & \speedup{1.0} & \speedup{1.5} & \speedup{2.0} \\
		
		\midrule
		\multicolumn{9}{c}{\textit{Group size $T \! = \! 8$ \, (theoretical speedup $\nicefrac{T^2}{2(T-1)} \! = \! \bm{4.57\times}$)}} \\
		\midrule
		Standard Linear & 4.194 & 8.590  & \textbf{0.54} & 1.90 & 7.25 & 28.95 & 120.23 & 570.75 \\
		Naive EQ-Linear & \textbf{0.524} & 8.590 & 0.55 & \textbf{1.89} & 7.22 & 29.19 & 120.99 & 574.21 \\
		\rowcolor{gray!10}
		\textbf{Flash EQ-Linear (PyTorch)} & \textbf{0.524} & \textbf{1.906}  & 1.01 & 3.12 & \textbf{7.01} & \textbf{16.91} & \textbf{44.94} & \textbf{138.08} \\
		\rowcolor{gray!10}
		\textit{Speedup/Reduction $\uparrow$} & \textbf{87.5\%} & \textbf{77.8\%} & \slowdown{0.5} & \slowdown{0.6} & \speedup{1.0} & \speedup{1.7} & \speedup{2.7} & \speedup{4.1} \\
		\bottomrule
	\end{tabular}
	\vspace{-2mm}
\end{table*}

\paragraph{Scaling with the group size $T$.}
Theorem~\ref{theorem:MACs_Flash_EQ-Linear} predicts that the speedup of Flash EQ-Linear grows with the group size $T$.
To verify this, Table~\ref{tab:flash-eqlinear-T-scaling} extends the operator-level comparison of Table~\ref{tab:flash-eqlinear-forward} from the $\mathrm{p}4$ group ($T \! = \! 4$) to $T \! \in \! \{2, 4, 8\}$, using the same PyTorch implementation strategy for every $T$.
Standard Linear and Naive EQ-Linear both compute a dense $C \! \times \! C$ projection with $NC^2$ MACs, whereas Flash EQ-Linear requires only $2(T \! - \! 1)NC^2/T^2$ MACs: two real matrix multiplications at the zero and Nyquist frequencies and $T/2 \! - \! 1$ complex ones, each mapping $\nicefrac{C}{T}$ input channels to $\nicefrac{C}{T}$ output channels.
The theoretical speedup $\nicefrac{T^2}{2(T-1)}$ therefore rises from $2.00\times$ ($T \! = \! 2$) to $2.67\times$ ($T \! = \! 4$) and $4.57\times$ ($T \! = \! 8$).
At a fixed total width $C \! = \! 2048$, the resulting FLOPs reductions (49.9\%, 62.4\%, and 77.8\%) lie within $0.4$ percentage points of the ideal $1 \! - \! 2(T \! - \! 1)/T^2$, since the group-wise DFT and IDFT contribute only lower-order cost.
The measured latency follows the same trend.
At $\nicefrac{C}{T} \! = \! 2048$, Flash EQ-Linear (PyTorch) is $1.6\times$, $2.0\times$, and $4.1\times$ faster than \texttt{F.linear} for $T \! = \! 2, 4, 8$, respectively, and for every $\nicefrac{C}{T} \! \geq \! 128$ the speedup increases monotonically with $T$ (e.g., $1.2\times \! \to \! 1.5\times \! \to \! 2.7\times$ at $\nicefrac{C}{T} \! = \! 1024$).
The channel width at which Flash EQ-Linear first outperforms \texttt{F.linear} also halves with each doubling of $T$ ($\nicefrac{C}{T} \! = \! 1024$, $512$, and $256$): at a fixed $\nicefrac{C}{T}$, the dense cost grows as $T^2$, whereas the memory traffic, layout transformations, and kernel-launch overhead of the unfused implementation grow only linearly in $T$.
The numerical accuracy is likewise unaffected by $T$.
In FP32, the Rel $L_2$ error remains below $10^{-6}$ for every $T$, the magnitude expected from floating-point round-off as in the numerical-equivalence test above.

\begin{table*}[t]
	\centering
		\caption{
		Comparison of standard ViT, naive EQ-ViT, Quick ViTs, and Flash EQ-ViT on ImageNet-100 in FP32.
		Within each scale, the best Top-1 accuracy, latency, and throughput are shown in \textbf{bold} and the second best are \underline{underlined}; parameter counts and FLOPs are reported as budgets and are not ranked.
		\textit{Speedup ratios} (\textcolor{green!50!black}{green}) are measured against standard ViT.
	}
	\vspace{-3mm}
	\label{tab:flash_Octic_more}
	\setlength{\tabcolsep}{2pt}
	\newcommand{\flashrow}{\rowcolor{gray!10}}
	\renewcommand{\arraystretch}{1.12}
	\resizebox{\textwidth}{!}{
		\begin{tabular}{l c c c c c}
			\toprule
			{Method}
			& {\makecell{\#Param. (M)$\downarrow$}}
			& {\makecell{FLOPs (G)$\downarrow$}}
			& {\makecell{Top-1 (\%)$\uparrow$}}
			& {\makecell{Latency (ms)$\downarrow$}}
			& {\makecell{Throughput (imgs/s)$\uparrow$}}
			\\
			\midrule
			ViT-T~\citep{dosovitskiy2021vit} & 21.7 & 9.2 & 76.5 & 0.358 & 2796 \\
			Naive EQ-ViT-T~\citep{fu2026vanilla} & 5.4 & 9.2 & \textbf{77.3} & 0.359 & 2785 \\
			Quick ViT-$\mathcal{D}_8$-T~\citep{nordstrom2026octic} & 2.8 & 1.7 & 71.6 & \underline{0.304} & \underline{3294}\,{\scriptsize\textcolor{green!50!black}{(1.2$\times$)}} \\
			Quick ViT-$\mathcal{H}_8$-T~\citep{nordstrom2026octic} & 4.3 & 2.3 & 76.0 & 0.305 & 3276\,{\scriptsize\textcolor{green!50!black}{(1.2$\times$)}} \\
			Quick ViT-$\mathcal{I}_8$-T~\citep{nordstrom2026octic} & 4.5 & 2.3 & \underline{76.9} & 0.307 & 3257\,{\scriptsize\textcolor{green!50!black}{(1.2$\times$)}} \\
			\flashrow Flash EQ-ViT-T & 5.4 & 4.0 & \textbf{77.3} & \textbf{0.249} & \textbf{4012}\,{\scriptsize\textcolor{green!50!black}{(1.4$\times$)}} \\
			\midrule
			ViT-S~\citep{dosovitskiy2021vit} & 33.8 & 14.1 & 76.5 & 0.603 & 1658 \\
			Naive EQ-ViT-S~\citep{fu2026vanilla} & 8.4 & 14.1 & 78.2 & 0.604 & 1656 \\
			Quick ViT-$\mathcal{D}_8$-S~\citep{nordstrom2026octic} & 4.3 & 2.6 & 76.3 & \underline{0.445} & \underline{2249}\,{\scriptsize\textcolor{green!50!black}{(1.4$\times$)}} \\
			Quick ViT-$\mathcal{H}_8$-S~\citep{nordstrom2026octic} & 6.7 & 3.5 & \underline{78.3} & 0.451 & 2220\,{\scriptsize\textcolor{green!50!black}{(1.3$\times$)}} \\
			Quick ViT-$\mathcal{I}_8$-S~\citep{nordstrom2026octic} & 6.9 & 3.6 & \textbf{78.7} & 0.454 & 2203\,{\scriptsize\textcolor{green!50!black}{(1.3$\times$)}} \\
			\flashrow Flash EQ-ViT-S & 8.4 & 6.0 & 78.2 & \textbf{0.429} & \textbf{2332}\,{\scriptsize\textcolor{green!50!black}{(1.4$\times$)}} \\
			\midrule
			ViT-B~\citep{dosovitskiy2021vit} & 85.9 & 35.1 & 77.1 & 1.135 & 881 \\
			Naive EQ-ViT-B~\citep{fu2026vanilla} & 21.4 & 35.1 & \textbf{80.1} & 1.128 & 886 \\
			Quick ViT-$\mathcal{D}_8$-B~\citep{nordstrom2026octic} & 10.8 & 6.5 & 74.7 & \textbf{0.720} & \textbf{1389}\,{\scriptsize\textcolor{green!50!black}{(1.6$\times$)}} \\
			Quick ViT-$\mathcal{H}_8$-B~\citep{nordstrom2026octic} & 29.5 & 13.3 & \underline{77.3} & 0.811 & 1233\,{\scriptsize\textcolor{green!50!black}{(1.4$\times$)}} \\
			Quick ViT-$\mathcal{I}_8$-B~\citep{nordstrom2026octic} & 29.9 & 13.5 & 77.0 & 0.815 & 1227\,{\scriptsize\textcolor{green!50!black}{(1.4$\times$)}} \\
			\flashrow Flash EQ-ViT-B & 21.4 & 14.3 & \textbf{80.1} & \underline{0.764} & \underline{1309}\,{\scriptsize\textcolor{green!50!black}{(1.5$\times$)}} \\
			\bottomrule
		\end{tabular}
	}
	\vspace{-2mm}
\end{table*}

\textbf{Comparison with Quick ViTs.}
Quick ViTs~\citep{nordstrom2026octic} build equivariance into the architecture by parameterizing features in the isotypic basis of the octic group $\mathrm{D}_8$, which comprises $90^\circ$ rotations and reflections.
Because this group is twice as large as ours ($|\mathrm{D}_8|\!=\!8$ versus $T\!=\!4$), its equivariant linear layers admit $5.33\times$ fewer FLOPs and $8\times$ fewer parameters than a dense layer---roughly twice the arithmetic reduction available to our formulation ($2.67\times$).
These larger theoretical savings make Quick ViTs a useful comparison for assessing how effectively our fused implementation translates arithmetic reductions into practical acceleration.
We compare Flash EQ-ViT with three Quick ViT configurations, denoted $\mathcal{D}_8$, $\mathcal{I}_8$, and $\mathcal{H}_8$.
The $\mathcal{D}_8$ variant retains octic-equivariant blocks throughout the backbone, whereas the hybrid $\mathcal{I}_8$ and $\mathcal{H}_8$ variants combine equivariant and standard blocks through tokenwise invariantization and relaxation of equivariance, respectively.
We adjust the proportion of equivariant and standard blocks in $\mathcal{I}_8$ and $\mathcal{H}_8$ to obtain comparable parameter budgets and train them on ImageNet-100 from scratch.
All measurements are obtained under identical experimental settings on a single NVIDIA RTX~4090 GPU and averaged over $200$ runs.

As shown in Table~\ref{tab:flash_Octic_more}, Flash EQ-ViT is generally faster than $\mathcal{H}_8$ and $\mathcal{I}_8$ at comparable accuracy, and substantially more accurate than $\mathcal{D}_8$.
Quick ViTs are equivariant to the larger octic group $\mathrm{D}_8$ and thus admit a larger arithmetic reduction ($5.33\times$ vs.\ $2.67\times$).
Even so, at the Tiny scale Flash EQ-ViT is $\bm{1.22\times}$ faster than the fastest Quick ViT variant while being more accurate, which we attribute to our fused CUDA kernels.
As width grows, the larger arithmetic reduction of $\mathrm{D}_8$ increasingly offsets this kernel-level advantage: at the Base scale, Quick ViT-$\mathcal{D}_8$ is $1.06\times$ faster but $5.4$ points less accurate (Table~\ref{tab:flash_Octic_more}).
Notably, Quick ViTs are new architectures trained from scratch, whereas Flash EQ-Linear accelerates existing models with no retraining and no accuracy change.

\end{document}

%% file: math_commands.tex
\usepackage{amsmath,amsfonts,bm}

\def\eqref#1{equation~\ref{#1}}

\def\1{\bm{1}}

\DeclareMathAlphabet{\mathsfit}{\encodingdefault}{\sfdefault}{m}{sl}
\SetMathAlphabet{\mathsfit}{bold}{\encodingdefault}{\sfdefault}{bx}{n}



%% file: references.bib
@inproceedings{dosovitskiy2021vit,
	title     = {An Image is Worth 16x16 Words: Transformers for Image
	Recognition at Scale},
	author    = {Dosovitskiy, Alexey and Beyer, Lucas and Kolesnikov,
	Alexander and Weissenborn, Dirk and Zhai, Xiaohua and
	Unterthiner, Thomas and Dehghani, Mostafa and Minderer,
	Matthias and Heigold, Georg and Gelly, Sylvain and
	Uszkoreit, Jakob and Houlsby, Neil},
	booktitle = {International Conference on Learning Representations},
	year      = {2021}
}

@inproceedings{liu2021swin,
	title     = {Swin Transformer: Hierarchical Vision Transformer
	using Shifted Windows},
	author    = {Liu, Ze and Lin, Yutong and Cao, Yue and Hu, Han and
	Wei, Yixuan and Zhang, Zheng and Lin, Stephen and Guo,
	Baining},
	booktitle = {Proceedings of the IEEE/CVF International Conference
	on Computer Vision},
	pages     = {10012--10022},
	year      = {2021}
}

@inproceedings{cohen2016group,
	title={Group Equivariant Convolutional Networks},
	author={Cohen, Taco S. and Welling, Max},
	booktitle={International Conference on Machine Learning},
	pages={2990--2999},
	year={2016}
}

@inproceedings{weiler2019general,
	title={General {E}(2)-Equivariant Steerable {CNN}s},
	author={Weiler, Maurice and Cesa, Gabriele},
	booktitle={Advances in Neural Information Processing Systems},
	year={2019}
}

@article{xie2022fourier,
	title={Fourier series expansion based filter parametrization for equivariant convolutions},
	author={Xie, Qi and Zhao, Qian and Xu, Zongben and Meng, Deyu},
	journal={IEEE Transactions on Pattern Analysis and Machine Intelligence},
	volume={45},
	number={4},
	pages={4537--4551},
	year={2022},
	publisher={IEEE}
}

@article{xie2025rotation,
	title={Rotation equivariant arbitrary-scale image super-resolution},
	author={Xie, Qi and Fu, Jiahong and Xu, Zongben and Meng, Deyu},
	journal={IEEE Transactions on Pattern Analysis and Machine Intelligence},
	year={2025},
	publisher={IEEE}
}

@article{he2021efficient,
	title={Efficient equivariant network},
	author={He, Lingshen and Chen, Yuxuan and Dong, Yiming and Wang, Yisen and Lin, Zhouchen and others},
	journal={Advances in Neural Information Processing Systems},
	volume={34},
	pages={5290--5302},
	year={2021}
}

@article{fu2026vanilla,
	title={Vanilla Group Equivariant Vision Transformer: Simple and Effective},
	author={Fu, Jiahong and Xie, Qi and Meng, Deyu and Xu, Zongben},
	journal={arXiv preprint arXiv:2602.08047},
	year={2026}
}

@article{zhao2026rotation,
	title={Rotation Equivariant Mamba for Vision Tasks},
	author={Zhao, Zhongchen and Xie, Qi and Huang, Keyu and Zhang, Lei and Meng, Deyu and Xu, Zongben},
	journal={arXiv preprint arXiv:2603.09138},
	year={2026}
}

@article{cheng2017survey,
	title={A Survey of Model Compression and Acceleration for Deep Neural Networks},
	author={Cheng, Yu and Wang, Duo and Zhou, Pan and Zhang, Tao},
	journal={IEEE Signal Processing Magazine},
	volume={35},
	number={1},
	pages={126--136},
	year={2018}
}

@inproceedings{han2015deep,
	title={Deep Compression: Compressing Deep Neural Networks with Pruning, Trained Quantization and Huffman Coding},
	author={Han, Song and Mao, Huizi and Dally, William J.},
	booktitle={International Conference on Learning Representations},
	year={2016}
}

@article{hinton2015distilling,
	title={Distilling the Knowledge in a Neural Network},
	author={Hinton, Geoffrey and Vinyals, Oriol and Dean, Jeff},
	journal={arXiv preprint arXiv:1503.02531},
	year={2015}
}

@article{mathieu2013fast,
	title={Fast training of convolutional networks through ffts},
	author={Mathieu, Michael and Henaff, Mikael and LeCun, Yann},
	journal={arXiv preprint arXiv:1312.5851},
	year={2013}
}

@article{vasilache2014fast,
	title={Fast convolutional nets with fbfft: A GPU performance evaluation},
	author={Vasilache, Nicolas and Johnson, Jeff and Mathieu, Michael and Chintala, Soumith and Piantino, Serkan and LeCun, Yann},
	journal={arXiv preprint arXiv:1412.7580},
	year={2014}
}

@article{cooley1965algorithm,
	title={An algorithm for the machine calculation of complex Fourier series},
	author={Cooley, James W and Tukey, John W},
	journal={Mathematics of computation},
	volume={19},
	number={90},
	pages={297--301},
	year={1965},
	publisher={JSTOR}
}

@article{cohen2016steerable,
	title={Steerable cnns},
	author={Cohen, Taco S and Welling, Max},
	journal={arXiv preprint arXiv:1612.08498},
	year={2016}
}

@inproceedings{weiler2018learning,
	title={Learning steerable filters for rotation equivariant cnns},
	author={Weiler, Maurice and Hamprecht, Fred A and Storath, Martin},
	booktitle={Proceedings of the IEEE Conference on Computer Vision and Pattern Recognition},
	pages={849--858},
	year={2018}
}

@inproceedings{shen2020pdoeconvs,
	title={Pdo-econvs: Partial differential operator based equivariant convolutions},
	author={Shen, Zhengyang and He, Lingshen and Lin, Zhouchen and Ma, Jinwen},
	booktitle={International Conference on Machine Learning},
	pages={8697--8706},
	year={2020},
	organization={PMLR}
}

@inproceedings{shen2021pdo,
	title={PDO-eS2CNNs: Partial differential operator based equivariant spherical CNNs},
	author={Shen, Zhengyang and Shen, Tiancheng and Lin, Zhouchen and Ma, Jinwen},
	booktitle={Proceedings of the AAAI conference on artificial intelligence},
	volume={35},
	number={11},
	pages={9585--9593},
	year={2021}
}

@inproceedings{kondor2018generalization,
	title={On the generalization of equivariance and convolution in neural networks to the action of compact groups},
	author={Kondor, Risi and Trivedi, Shubhendu},
	booktitle={International Conference on Machine Learning},
	pages={2747--2755},
	year={2018},
	organization={PMLR}
}

@inproceedings{ravanbakhsh2017equivariance,
	title={Equivariance through parameter-sharing},
	author={Ravanbakhsh, Siamak and Schneider, Jeff and P{\'o}czos, Barnab{\'a}s},
	booktitle={International Conference on Machine Learning},
	pages={2892--2901},
	year={2017},
	organization={PMLR}
}

@article{chetlur2014cudnn,
	title={cuDNN: Efficient primitives for deep learning},
	author={Chetlur, Sharan and Woolley, Cliff and Vandermersch, Philippe and Cohen, Jonathan and Tran, John and Catanzaro, Bryan and Shelhamer, Evan},
	journal={arXiv preprint arXiv:1410.0759},
	year={2014}
}

@inproceedings{dao2022flashattention,
	title={FlashAttention: Fast and memory-efficient exact attention with IO-awareness},
	author={Dao, Tri and Fu, Daniel Y and Ermon, Stefano and Rudra, Atri and R{\'e}, Christopher},
	booktitle={Advances in Neural Information Processing Systems},
	volume={35},
	pages={16344--16359},
	year={2022}
}

@article{williams2009roofline,
	title={Roofline: An insightful visual performance model for multicore architectures},
	author={Williams, Samuel and Waterman, Andrew and Patterson, David},
	journal={Communications of the ACM},
	volume={52},
	number={4},
	pages={65--76},
	year={2009},
	publisher={ACM}
}

@inproceedings{deng2009imagenet,
	title={ImageNet: A large-scale hierarchical image database},
	author={Deng, Jia and Dong, Wei and Socher, Richard and Li, Li-Jia and Li, Kai and Fei-Fei, Li},
	booktitle={IEEE Conference on Computer Vision and Pattern Recognition},
	pages={248--255},
	year={2009}
}

@inproceedings{gerken2022equivariance,
	title={Equivariance versus Augmentation for Spherical Images},
	author={Gerken, Jan and Carlsson, Oscar and Linander, Hampus and Ohlsson, Fredrik and Petersson, Christoffer and Persson, Daniel},
	booktitle={Proceedings of the 39th International Conference on Machine Learning},
	pages={7404--7421},
	volume={162},
	series={Proceedings of Machine Learning Research},
	publisher={PMLR},
	year={2022}
}

@inproceedings{finzi2021practical,
	title={A Practical Method for Constructing Equivariant Multilayer Perceptrons for Arbitrary Matrix Groups},
	author={Finzi, Marc and Welling, Max and Wilson, Andrew Gordon},
	booktitle={Proceedings of the 38th International Conference on Machine Learning},
	series={Proceedings of Machine Learning Research},
	volume={139},
	pages={3318--3328},
	publisher={PMLR},
	year={2021}
}

@inproceedings{paszke2019pytorch,
	title={PyTorch: An Imperative Style, High-Performance Deep Learning Library},
	author={Paszke, Adam and Gross, Sam and Massa, Francisco and Lerer, Adam
	and Bradbury, James and Chanan, Gregory and Killeen, Trevor
	and Lin, Zeming and Gimelshein, Natalia and Antiga, Luca
	and Desmaison, Alban and Kopf, Andreas and Yang, Edward
	and DeVito, Zachary and Raison, Martin and Tejani, Alykhan
	and Chilamkurthy, Sasank and Steiner, Benoit and Fang, Lu
	and Bai, Junjie and Chintala, Soumith},
	booktitle={Advances in Neural Information Processing Systems},
	volume={32},
	pages={8024--8035},
	year={2019}
}

@book{oppenheim2009discrete,
	title={Discrete-Time Signal Processing},
	author={Oppenheim, Alan V and Schafer, Ronald W},
	edition={3},
	publisher={Pearson},
	year={2009}
}

@inproceedings{nordstrom2026octic,
	title={Octic Vision Transformers: Quicker ViTs Through Equivariance}, 
	author={David Nordström and Johan Edstedt and Fredrik Kahl and Georg Bökman},
	booktitle={Proceedings of the European Conference on Computer Vision (ECCV)},
	year={2026}
}

@inproceedings{bokman2025flopping,
	title     = {Flopping for {FLOP}s: Leveraging Equivariance for Computational Efficiency},
	author    = {B{\"o}kman, Georg and Nordstr{\"o}m, David and Kahl, Fredrik},
	booktitle = {Proceedings of the 42nd International Conference on Machine Learning},
	series    = {Proceedings of Machine Learning Research},
	volume    = {267},
	pages     = {4823--4838},
	publisher = {PMLR},
	year      = {2025}
}

@book{hwu2022programming,
	title     = {Programming Massively Parallel Processors: A Hands-on Approach},
	author    = {Hwu, Wen-mei W. and Kirk, David B. and El Hajj, Izzat},
	edition   = {Fourth},
	publisher = {Morgan Kaufmann},
	address   = {Cambridge, MA},
	year      = {2022},
	isbn      = {978-0-323-91231-0}
}

@book{serre1977linear,
	title     = {Linear Representations of Finite Groups},
	author    = {Serre, Jean-Pierre},
	series    = {Graduate Texts in Mathematics},
	volume    = {42},
	publisher = {Springer},
	year      = {1977}
}

@inproceedings{romero2021group,
	title     = {Group Equivariant Stand-Alone Self-Attention For Vision},
	author    = {Romero, David W. and Cordonnier, Jean-Baptiste},
	booktitle = {International Conference on Learning Representations},
	year      = {2021},
	url       = {https://openreview.net/forum?id=JkfYjnOEo6M}
}

@inproceedings{xu2023e2vit,
	title     = {{$E(2)$}-Equivariant Vision Transformer},
	author    = {Xu, Renjun and Yang, Kaifan and Liu, Ke and He, Fengxiang},
	booktitle = {Proceedings of the Thirty-Ninth Conference on Uncertainty in Artificial Intelligence},
	series    = {Proceedings of Machine Learning Research},
	volume    = {216},
	pages     = {2356--2366},
	publisher = {PMLR},
	year      = {2023},
	url       = {https://proceedings.mlr.press/v216/xu23b.html}
}
